\documentclass[10pt,twocolumn,letterpaper]{article}

\usepackage{cvpr}              
\makeatletter
\@namedef{ver@everyshi.sty}{}
\makeatother

\usepackage{graphicx}
\usepackage{amsmath}
\usepackage{amssymb}
\usepackage{booktabs}
\usepackage{makecell}
\usepackage{arydshln}
\usepackage{enumitem}
\usepackage{array,multirow}
\usepackage{float}
\usepackage{pifont}
\usepackage{adjustbox}
\usepackage{tabularray}
\usepackage{tikz}

\usepackage{enumitem}
\usepackage[accsupp]{axessibility}

\usepackage[pagebackref,breaklinks,colorlinks]{hyperref}

\usepackage[capitalize]{cleveref}
\crefname{section}{Sec.}{Secs.}
\Crefname{section}{Section}{Sections}
\Crefname{table}{Table}{Tables}
\crefname{table}{Tab.}{Tabs.}

\def\cvprPaperID{3417} 
\def\confName{CVPR}
\def\confYear{2023}

\newcommand{\cmark}{\ding{51}}%
\newcommand{\xmark}{\ding{55}}%

\begin{document}

\title{AutoFocusFormer: Image Segmentation off the Grid}

\author{
Chen Ziwen$^{1\thanks{Work done while Chen Ziwen was an intern at Apple Inc.}}$, 
Kaushik Patnaik$^{2}$,
Shuangfei Zhai$^{2}$,
Alvin Wan$^{2}$ \\
Zhile Ren$^{2}$,
Alex Schwing$^{2}$,
Alex Colburn$^{2}$, 
Li Fuxin$^{1,2}$ \\
$^1$Oregon State University, $^2$Apple Inc. \\
{\tt \small \{chenziw, lif\}@oregonstate.edu}\\
{\tt \small \{kaushik\_patnaik,szhai,alvinwan,zhile\_ren,aschwing,alexcolburn,fli26\}@apple.com}
}
\maketitle

\begin{abstract}
   Real world images often have highly imbalanced content density. Some areas are very uniform, e.g., large patches of blue sky, while other areas are scattered with many small objects. Yet, the commonly used successive grid downsampling strategy in convolutional deep networks treats all areas equally. Hence, small objects are represented in very few spatial locations, leading to worse results in tasks such as segmentation. 
   Intuitively, retaining more pixels representing small objects during downsampling helps to preserve important information. 
   To achieve this, we propose AutoFocusFormer (AFF), a local-attention transformer image recognition backbone, which performs adaptive downsampling by learning to retain the most important pixels for the task.
   Since adaptive downsampling generates a set of pixels irregularly distributed on the image plane, we abandon the classic grid structure.
   Instead, we develop a novel point-based local attention block, facilitated by a balanced clustering module and a learnable neighborhood merging module, which yields representations for
   our point-based versions of state-of-the-art segmentation heads. 
   Experiments show that our AutoFocusFormer (AFF) improves significantly over baseline models of similar sizes.


\end{abstract}


\begin{figure}
    \centering
    \includegraphics[width=\linewidth]{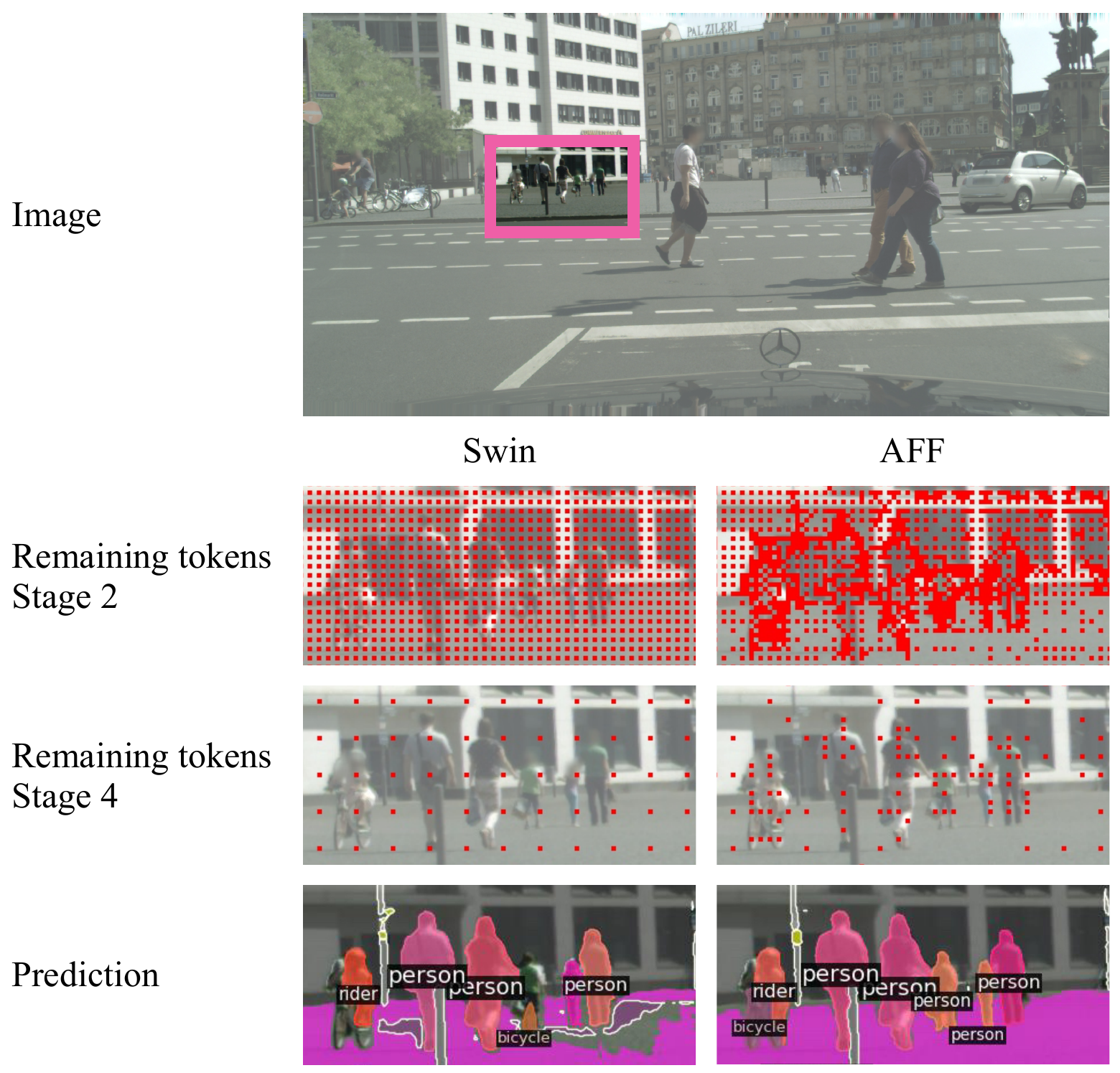}
    \vskip -0.1in
    \caption{Comparison between on-grid model Swin~\cite{swin} and off-grid model AFF. The red pixels indicate the locations of the remaining tokens. AFF downsamples non-uniformly, automatically focusing on more textured, important image regions, which lead to better performance on small objects in the scene. }
    \label{fig:city_first}
    \vskip -0.15in
\end{figure}

\section{Introduction}
\label{sec:intro}


Typical real-world images distribute content unevenly. Consider the photo of a typical outdoor scene in Fig.~\ref{fig:city_first}: Large swaths of the image contain textureless regions like the ground, while a few regions contain many small objects. 
Despite this, most computer vision neural networks distribute computation evenly across the image; every pixel, regardless of texture or importance, is processed with the same computational cost. Popular convolutional neural networks operate on regularly-arranged square patches. Although recent transformer architectures do not strictly depend on a grid structure, many transformer-based methods adopt grid-based techniques such as stride-16 convolutions~\cite{vit} and $7\times 7$ square windows for local attention~\cite{swin}.

\begin{figure*}[t]
    \centering
    \includegraphics[width=\linewidth]{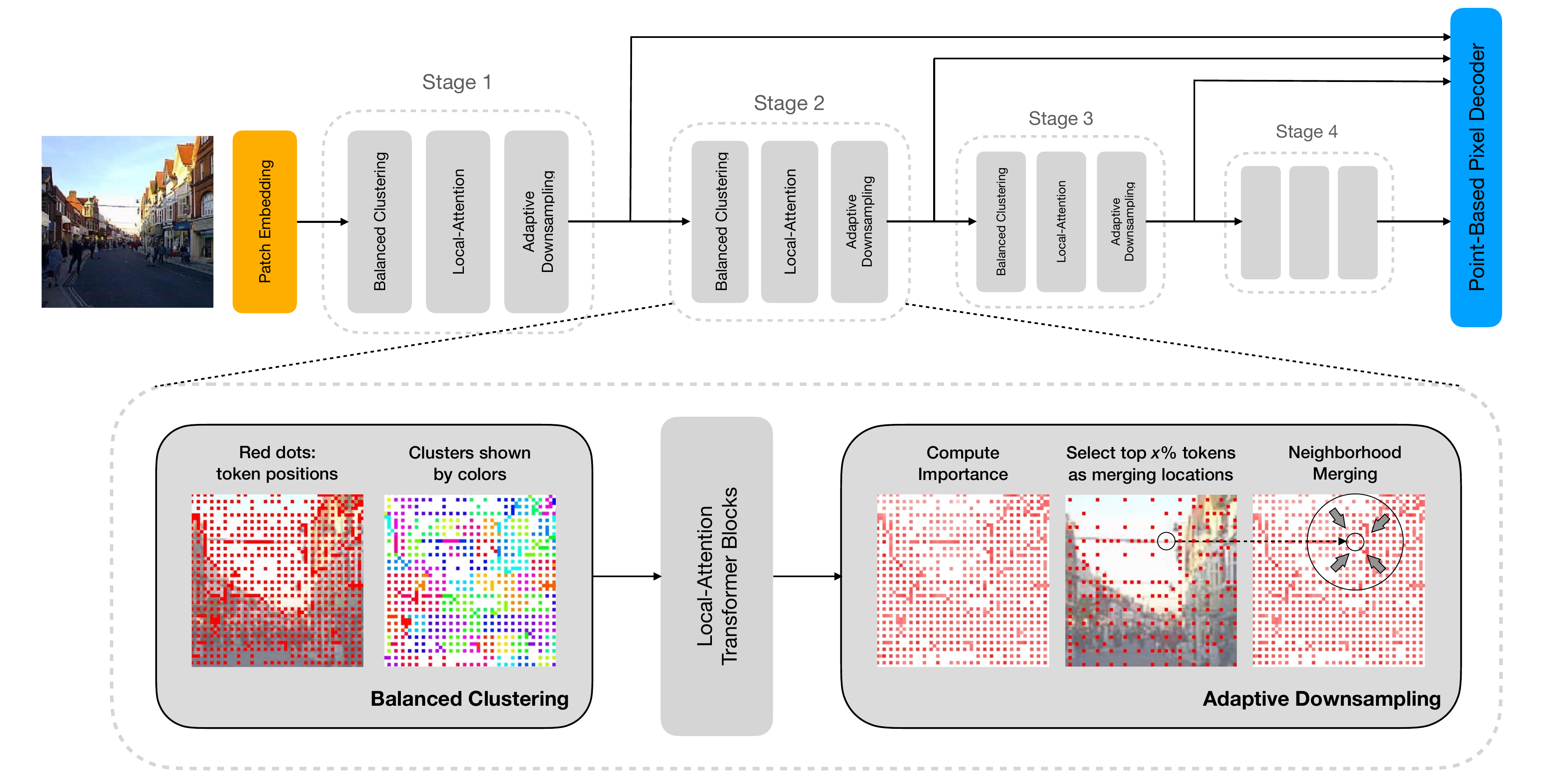}
    \vskip -0.1in
    \caption{The network architecture of AutoFocusFormer. The model consists of four stages, each stage processing a successively downsampled set of tokens. Within each stage, tokens first go through balanced clustering, then attend to the tokens in their local neighborhoods defined by the nearby clusters in the following local-attention blocks, and finally adaptively merge into the set of downsampled output tokens with weights modulated by the learnable importance scores.}
    \label{fig:flowchart}
    \vskip -0.15in
\end{figure*}

Despite its popularity, uniform downsampling is less effective for tasks that require pixel-level details such as segmentation. Here, uniform downsampling unfortunately makes tiny objects even tinier -- possibly dropping needed, pixel-level information. To combat this, many techniques increase the input resolution~\cite{sniper, du2021simple} to obtain better segmentation performance. This intuitively helps, as larger input will lead to higher resolution after downsampling. However, increasing input resolution is costly in memory and computation, as this brute-force bandaid neglects the underlying issue -- namely, uniform downsampling. Some prior works amend this by irregularly sampling points in the segmentation decoder~\cite{kirillov2020pointrend}, but by still relying on a uniformly-downsampled convolutional encoder, these techniques remain susceptible to the pitfalls of uniform downsampling.

To address this concern, we need solutions that enable computer vision models to allocate computation non-uniformly across each image. In particular, we need a downsampling strategy that retains important details, while more aggressively summarizing texture-less regions such as sky or road. However, non-uniform downsampling breaks from the grid structure that existing architectures rely on. Prior work on adaptive downsampling~\cite{dynamicvit,notall,adaptivetoken} addresses this by simply using global attention, but global attention does not scale to resolutions much higher than that of ImageNet, such as those required for segmentation tasks.


To satisfy this need for adaptive, scalable downsampling strategies, we propose \textit{AutoFocusFormer (AFF)}. To our knowledge, AFF is the first \textbf{end-to-end segmentation network with successive adaptive downsampling stages}. To scale to higher resolutions required in segmentation tasks, AFF employs local attention blocks. In order to define local attention neighborhoods among irregularly sampled tokens, we develop a novel balanced clustering algorithm which employs space-filling curves to group irregular locations into neighborhoods. We also propose a novel adaptive downsampling module that learns the importance of different image locations through a differentiable neighborhood merging process 
(Fig.~\ref{fig:merge}). Finally, we modify state-of-the-art segmentation heads so that they can be applied on the irregular-spaced representations our backbone generates.


Our AutoFocusFormer attains state-of-the-art performance with less computational cost across major segmentation tasks, with especially strong results when using smaller models. Furthermore, by moving away from the grid structure, our downsampling strategy can support a larger range of computational budget by retaining any number of tokens, rather than operating only at rates of $1/4$, $1/16$ etc.


To summarize, our contributions are: 
\begin{itemize}[noitemsep,topsep=0pt,parsep=0pt,partopsep=0pt]
    \item To our knowledge, we introduce the first end-to-end segmentation network with successive adaptive downsampling stages and with flexible downsampling rates.
    \item To facilitate a local attention transformer on irregularly spaced tokens, we propose a novel balanced clustering algorithm to group tokens into neighborhoods.  
    We also propose a neighborhood merging module that enables  end-to-end learning of adaptive downsampling.
    \item We adapt state-of-the-art decoders such as deformable  DETR~\cite{deformable}, Mask2Former~\cite{mask2} and HCFormer~\cite{hcformer} to operate on irregularly spaced sets of tokens.
    \item Results show that our approach achieves  state-of-the-art for both image classification and segmentation with fewer FLOPs, and improves   significantly on the recognition of small objects in instance segmentation tasks. 
\end{itemize}



\section{Related Work}\label{sec:relwork}

\noindent\textbf{Transformer backbones for Vision.}
The seminal Vision Transformer (ViT)~\cite{vit} demonstrated that simple image patch encodings combined with self-attention enable impressive modeling capacity on image classification tasks. However, ViTs lack hierarchical feature maps which are critical for dense prediction tasks. Various improvements have since been proposed, e.g., MViT and PVT \cite{fan2021multiscale,wang2021pyramid} propose using feature hierarchies similar to those of standard convolutional networks. 
However, they still operate on global attention, which suffers from quadratic complexity w.r.t.\ the input size, hence struggling with high resolution inputs. One solution is to apply attention only on the low resolution feature maps, as in BoT~\cite{srinivas2021bottleneck} and LeViT~\cite{levit}. Other approaches modify the attention operation. SegFormer\cite{segformer} computes attention in earlier layers on a concatenation of many tokens.  PoolFormer and AFT\cite{yu2022metaformer,zhai2021attention} replace attention with pooling based operations which reduces the quadratic complexity to linear. 
Swin Transformer and SASA~\cite{swin,sasa} replace the global attention with local attention, where attention is computed only in a small local window, which achieves better efficiency. 

\noindent\textbf{Clustering-based Attention.} Clustering based sparse attention has been proposed in language and vision~\cite{roy2021efficient,vyas2020fast,zheng2020end,wang2021cluster}. However, these works do not involve adaptive downsampling which is central to our method. Most works attempt to cluster learned features, whereas we cluster based on token locations. We have also studied clustering of features, but found it not to provide significant improvements, while adding significant complexity to the model. Clustering ideas have also been applied in decoders~\cite{yu2022k,hcformer} which differs from our work as we apply clustering in the encoder.



\noindent\textbf{Adaptive downsampling.}
There have been many attempts to combine adaptive downsampling with vision Transformers, such as AdaViT~\cite{adavit}, DynamicViT~\cite{dynamicvit}, A-ViT~\cite{avit}, etc. 
Dynamic Grained Encoder\cite{song2021dynamic} proposes to learn different grid downsampling rates for different regions. EViT\cite{notall} proposes to merge the ``uninformative'' patches into one. PS-ViT~\cite{psvit} proposes to learn offsets to the original grid patch locations. However, all of these solutions are still based on global attention. Most of them do not prune away ``uninformative'' patches during training, either due to the need for gradients to flow through those patches~\cite{dynamicvit} or the need for uniform size across the batch~\cite{avit}. Hence, they cannot scale to high-resolution segmentation tasks, but only focus on speeding up ImageNet classification.

How to make the adaptive downsampling module learnable has itself been a significant challenge. Some methods turn to heuristics such as the attention values~\cite{adaptivetoken}; some turn to policy gradient~\cite{iared}; some turn to the Gumbel-Softmax trick~\cite{gumbel} to obtain a differentiable binary mask~\cite{adavit, dynamicvit}. To directly obtain gradients from the task loss, merging tokens~\cite{tokenlearner, notall} seems to be a more natural strategy than deleting tokens. We develop a novel neighborhood merging module, adaptively choosing the merging locations, providing gradients directly from the task loss to the ``importance scores'' of the tokens.
To our best knowledge, our work is the first end-to-end framework with local attention blocks that uses adaptive downsampling in multiple stages, and is scalable to high-resolution segmentation tasks.

\noindent\textbf{Point cloud networks.} Prior works that directly operate on a set of irregular points are mostly designed for 3D point clouds, such as PointNet++~\cite{pointnet++}, PointConv~\cite{pointconv} and Point Transformer~\cite{pointtransformer}. They often choose $k$-nearest-neighbors or an $\epsilon$-ball to find the neighborhood for each point. 
We make limited usage of PointConv in our decoder model to replace $3 \times 3$ convolution. 

\section{Method}\label{sec:method}
Our goal is to perform adaptive downsampling instead of the traditional grid-based downsampling. Specifically, we want to retain more 2D image locations in  ``informative'' areas (e.g., areas which depict cluttered small objects), and more succinctly summarize ``unimportant'' areas (e.g., a purely blue sky). Because the chosen locations  have uneven density, our model treats the locations as a set of irregularly-spaced tokens rather than a rectangular grid. 

As we mentioned earlier, global-attention transformers are computationally demanding for segmentation tasks. 
Thus, local transformer models are a natural choice for our goal due to their computation efficiency on larger images.  However, the regular grid local window mechanism in methods like Swin Transformer~\cite{swin} is not amenable to adaptive irregular downsampling, so we propose to cluster pixels and perform attention on clusters.

Specifically, our backbone model (Fig.~\ref{fig:flowchart}) starts with a patch embedding module (2 layers of $3\times 3$ convolution with stride 2), and then continues with several \textit{stages}. Each stage consists of: 1) a clustering algorithm (Section~\ref{sec:cluster}); 2) several local-attention transformer blocks (Section~\ref{sec:attention}); 3) the novel adaptive downsampling module (Section~\ref{sec:adads}).  Finally, task-specific heads are attached to the backbone for different tasks such as image classification, semantic and instance segmentation.

\begin{figure}
\includegraphics[width=1.0\columnwidth]{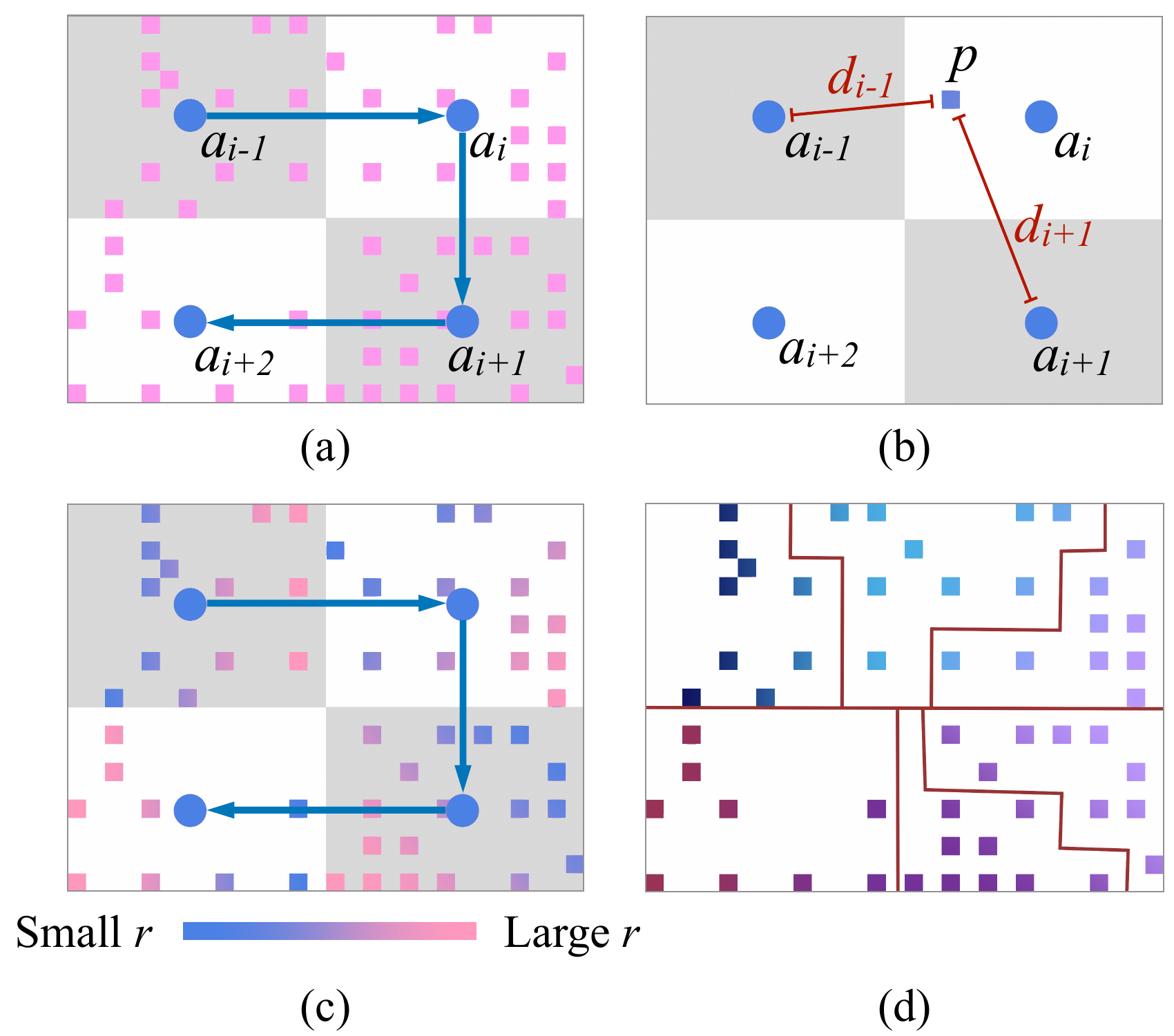} 
\vskip -0.05in
\caption{Illustration of the balanced clustering algorithm. (a) Tokens are quantized to space-filling anchors. A space-filling curve orders the anchors ($a_{i-1}, a_{i}, \cdots$). (b) For each token (e.g., token $p$ quantized to $a_i$), calculate the ratio of its distance to the ``previous" and the ``next" anchor ($r(p)=d_{i-1}(p) / d_{i+1}(p)$). (c) Order tokens quantized to the same anchor in ascending order of $r$. (d) Sort all tokens based on the anchor order and the local token order. Partition the sorted tokens into equal-sized clusters.}
\label{fig:cluster}
\vskip -0.2in
\end{figure}

\subsection{Clusters and Neighborhoods}\label{sec:cluster}

A local neighborhood on a 2D grid can be conveniently defined by slicing out a square window. In contrast, for an unordered set of points, the conventional approach to identify neighbors in a 3D point cloud~\cite{pointnet++,pointconv} relies on algorithms such as $k$-nearest-neighbors (kNN), which compute pairwise distances between points.

A naive kNN algorithm has a quadratic time complexity. Interestingly, many of the algorithms to speed up kNN (e.g.,~\cite{faiss}) involve a first step of $k$-means clustering on given points so as to reduce the search space for the neighbors. 
Inspired by this approach, we also use 
\textbf{clusters} in defining local neighborhoods, i.e., we  
divide the tokens into clusters, and define neighborhoods as entailing several 
nearby clusters. Unfortunately, traditional clustering algorithms such as $k$-means or locality-sensitive hashing do not directly fit our purpose. First, they require multiple iterations or rounds of hashing and are thus slow, compared to a forward pass in a deep net. More importantly, they result in clusters with different numbers of assigned points. Unevenly sized clusters results in tensors padded with zeros, which leads to wasted memory and time. 

Noting that we are only clustering on 2D positions but not higher dimensions, we propose to address the aforementioned concerns with a novel balanced clustering method.

\subsubsection{Balanced Clustering}
Instead of conventional iterative approaches, our method to obtain a perfectly balanced clustering is to arrange 
2D locations on the image 
into a 1D array, and then partition the array into groups of equal size. 
For this we consider space-filling curves~\cite{peano}, which arrange 2D locations on a 1D line while attempting to preserve the 2D distance measure~\cite{moon2001analysis}. Hence, points that are close on the line are also reasonably close in the 2D space. 
However, due to the conversion from 2D to 1D, it is impossible to completely preserve the 2D metric, and artifacts occur if we directly utilize space-filling curves to partition tokens into clusters. 
To partially alleviate this concern, we adopt a 2-stage process for the clustering. The idea is to utilize space-filling curves only at a coarse level to obtain an ordering for sparsely and regularly sampled 2D image locations. Then, tokens are ordered based on the 2D distances to these locations.

Concretely, we first divide the image into a coarse regular grid with the number of square patches being similar to the intended number of clusters. We refer to the center of each square patch in the grid  as a \textit{space-filling anchor}. A space-filling curve establishes an ordering among the anchors. 
Given this ordering, for a token with position $p\in \mathbb{R}^2$ belonging the anchor $a_i\in \mathbb{R}^2$, we can define its previous anchor $a_{i-1}$ and next anchor $a_{i+1}$. 
Then, we calculate the ratio $r$ of distances from token $p$ to the two anchors via
\begin{equation}
    r(p) = \dfrac{d_{i-1}(p)}{d_{i+1}(p)} = \dfrac{\|p-a_{i-1}\|_2}{\|p-a_{i+1}\|_2}
\end{equation}
for all tokens $p$.
Now, within each square patch, we order the tokens in ascending order of $r$, so that  tokens closer to the previous anchor are placed earlier. 
This procedure hence establishes an ordering of all the tokens. 
Finally, to obtain balanced clustering we simply partition the array into groups of equal size. Fig.~\ref{fig:cluster} illustrates the entire algorithm.

Because we can find the corresponding space-filling anchor for each token in $O(1)$ time by simply quantizing their coordinates, the overall time complexity of the clustering is no more than sorting all the token locations in their local patches once, which is negligible compared to the network time complexity because feature channels are not involved. Note that the clustering algorithm only needs to be performed once at the beginning of each stage, and the cluster information can be used for all attention blocks in the stage, as well as the downsampling module at the end.

Different from prior balanced clustering work~\cite{banerjee2006scalable}, this algorithm is not iterative and  results in perfectly balanced clusters. 
It also guarantees that each token belongs to a single cluster, different from RoutingTransformer~\cite{roy2021efficient}, where some tokens may not belong to any  cluster. 
However, note that our balanced clustering is only suitable for low-dimensional points. In early stages of this research, we explored clustering with embedded features instead of just 2D locations, but the performance difference was negligible. Hence, we decided to cluster on the locations only, which allowed us to utilize the proposed algorithm to generate perfectly balanced clusters. Please see the supplementary material for experiments validating our clustering approach.

\subsubsection{Neighborhoods from clusters} 
In order to encourage information flow across the whole image, it is important that attention is not limited to only locations within the same cluster. E.g., in Swin Transformers~\cite{swin}, shifting windows between consecutive layers allows pixels to attend to different neighbors in different layers. However, in our case, re-clustering every layer would add undesired computation. 
Hence, we 
opt to use 
smaller clusters and allow each token to attend to tokens from $R$ nearby clusters. To achieve this, we use a neighborhood size several times larger than the cluster size. This is beneficial because the neighborhoods are overlapping, guaranteeing  information exchange between the clusters. 

\subsection{Transformer Attention Blocks}\label{sec:attention}
At any stage, let $N$ be the number of tokens, $M$ be the number of neighbors for each token, $H$ be the number of heads, and $C$ be the number of feature channels. 
The local attention $A$ of one single head with relative position embedding is computed by having each token attend to all the other tokens in its neighborhood via
\begin{equation}
    A = \text{softmax}(QK^T+P),
\end{equation}
where 
$Q\in \mathbb{R}^{N\times C/H}, K\in \mathbb{R}^{M\times C/H}, A\in \mathbb{R}^{N\times M}, P\in \mathbb{R}^{N\times M}$ 
are query, key, attention and position embedding matrices with
\begin{equation}
    P_{i,j} = w(p_i-p_j). 
\end{equation}
Here, $p_i,p_j$ are the $(x,y)$ coordinates of two neighboring tokens, and $w(\cdot)$ is a function that returns a scalar position embedding for this head. For models with fixed-shape neighborhoods, the relative position embeddings can be stored in a matrix and read when needed. But for our model, the positions of the neighbors are unknown beforehand, and thus we need to project the difference of the coordinates to a position embedding via a learnable function. We implement $w$ as one fully-connected layer.

\begin{figure}
\centering
\includegraphics[width=0.8\columnwidth]{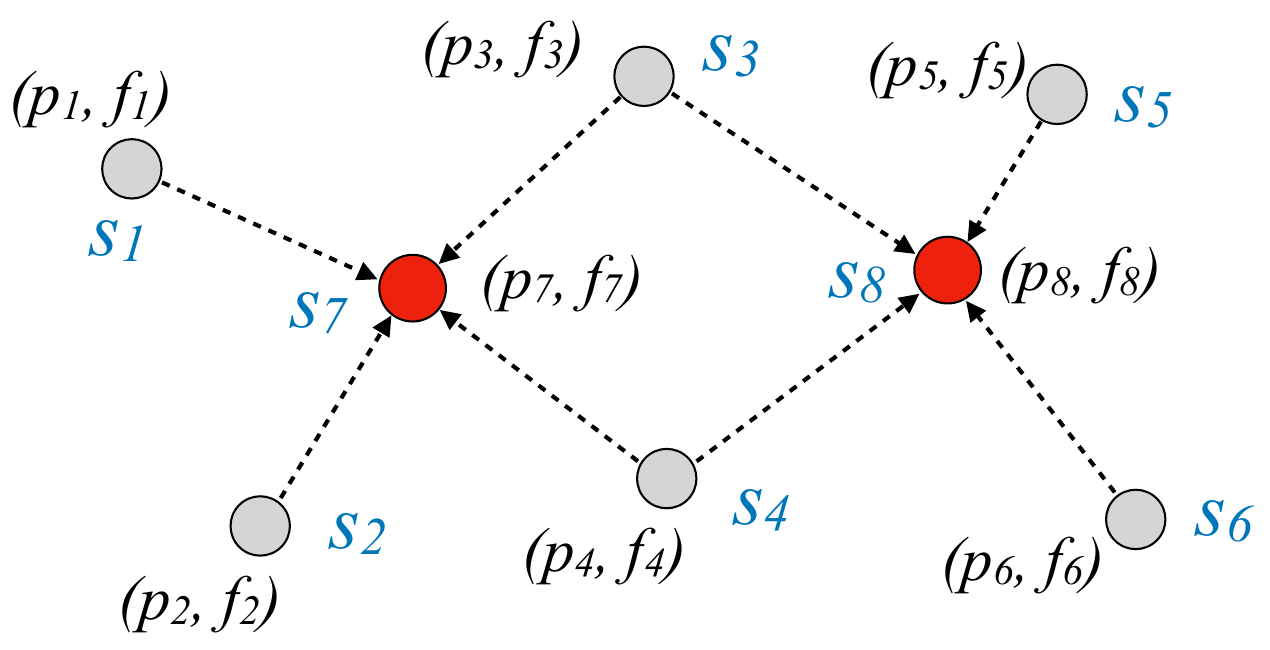}
\vskip -0.15in 
\caption{Illustration of the adaptive downsampling module. First, importance scores $s_i$ are calculated from the token features $f_i$. Then, tokens with the highest scores are selected as merging centers (e.g., $p_7$ and $p_8$). Finally, neighborhoods are merged with weights modulated by the importance scores $s_i$.}
\label{fig:merge}
\vskip -0.1in
\end{figure}

\subsection{Adaptive downsampling}\label{sec:adads}

At the end of each but the final stage, we employ a downsampling module. 
The downsampling module includes two components, a learnable scoring component for choosing the most 
\textit{important} tokens, and the \textit{neighborhood merging} step to merge the neighbors around selected tokens.

\subsubsection{Learnable Importance Score}
We learn to predict a scalar score $s_i$ for the $i$-th token to indicate its importance w.r.t.\ the task loss. 
A main difficulty in learning this score is that downsampling limits the number of selected tokens. If the task loss is only backpropagated to the selected tokens, tokens that were not selected might not receive any gradient. 
In order to backpropagate gradients to all the tokens, we propose to learn the importance score for each token in conjunction with the neighborhood merging process.

To illustrate this, assume we are merging a neighborhood around a particular token at location $p_c$. 
Similar to attention blocks, a neighborhood $\mathcal{N}(p_c)$ is obtained from its nearest $R$ clusters and the $i$-th neighboring token is denoted as a tuple $(p_i, f_i)$ with location $p_i = (x_i, y_i)$ and features $f_i \in \mathbb{R}^C$. 
In the neighborhood, merging is performed using a modification of a PointConv layer\cite{pointconv} as follows:  
  \vspace{-0.4cm}
\begin{equation}
    f_{\text{merged}}(p_c) \!=\! \text{vec} \left(\!\sum_{(p_i,f_i)\in \mathcal{N}(p_{\text{c}})} \hspace{-0.5cm}\overbrace{\sigma(l(f_i))}^\text{$s_i$}\cdot \mathbf{W}(p_i-p_{\text{c}})f_i^\top\right)U,
    \label{eq:merging}
\end{equation}
where $f_{\text{merged}}\in \mathbb{R}^{C^\prime}$ is the merged output with $C'$ outputs, $\text{vec}(\cdot)$ means vectorization, $\sigma(\cdot)$ is the sigmoid function, $l(\cdot)$ is a fully-connected layer predicting the scalar ``importance score", $\mathbf{W}(\cdot)$ is a multi-layer perceptron (MLP) with output shape $C_{\text{mid}}\times 1$, which creates different weighted combinations of the input features $f_i$ in the neighborhood. The weights are learned to be a function of the relative coordinates between the location of each neighboring token and the merging center location $p_c$. Finally, $U \in \mathbb{R}^{C_{\text{mid}} C \times C^\prime}$ is implemented as a fully-connected layer. 

Note, Eq.~\eqref{eq:merging} is similar to PointConv $f = \mathrm{vec}(\sum \mathbf{W}f^\top) U$, which was shown to be equivalent to one continuous convolution layer~\cite{pointconv}. We add $s_i = \sigma(l(f_i))$ to modulate this function, which allows the model 
%
to ``turn off'' the unimportant tokens during neighborhood merging, keeping their features from being utilized in the next stage. Thus, the score $s_i$ can be viewed as an indicator, denoting how important the model thinks a token is. This score can henceforth be used to select the merging centers. 

This formulation allows us to directly learn the score $s_i$ from the task loss without resorting to costly techniques such as a policy gradient used in prior work~\cite{iared}. 
Some previous works use Gumbel Softmax/Sigmoid~\cite{adavit} to obtain a differentiable binary mask, in order to 
keeping gradients flowing to the ``deleted" tokens after downsampling; since we adopt the merging strategy rather than deletion, we do not need such a hard binary mask.
We provide visualizations of the neighborhood merging process in Fig.~\ref{fig:merge}.

\subsubsection{Grid priors and Selection of Tokens to Sample}
\label{sec:grid_prior}

The final score used for selecting tokens as merging centers is $g_i + \alpha s_i$, a weighted sum between the learned score $s_i$ and a grid prior $g_i$, with a hyperparameter $\alpha$. While $s_i$ is only based on the feature vector, the grid prior helps the model to differentiate between similar tokens at different locations, thus facilitating the model to perform proper, uniform-stride downsampling in local texture-less regions.
Our grid prior $g_i$ mimics the behavior of the traditional grid downsampling -- alternatingly assigning 1 and 0 to the tokens. However, in stages where the tokens are already irregularly sampled, adding a regular grid prior is no longer reasonable. Hence, we advocate for an ``adaptive'' grid prior that takes the local density of the sampled tokens into account.

Specifically, the adaptive prior chooses a local grid level based on the local stride of the sampled tokens. For each token $p_i = (x_i, y_i)$, we assign its ``stride'' $t_i = 2^{ \lceil \log_2 \min_j \|p_i - p_j\|_1\rceil}$ to be the distance from the token to its nearest neighbor, rounded up to the nearest power of 2. For example, if the token is $3$ pixels apart from its nearest neighbor, it is assigned stride $t_i=4$. Then, we assign $g_i = 1$ if $(x_i \mod 2 t_i = 0) \wedge  (y_i \mod 2 t_i = 0)$. That is, we want to downsample the local stride from $t_i$ to $2t_i$. Hence, if $t_i = 1$, then $g_i = 1$ on alternating pixels, and if $t_i = 2$, then $g_i = 1$ every 4 pixels. 
Furthermore, we set the grid prior to infinity for tokens with ($x_i \mod 2^{j+1} = 0) \wedge (y_i \mod 2^{j+1} = 0)$ in the $j$-th stage.  We call these tokens ``reserved". We reserve these coarse-grid tokens to ensure the connectivity among remote regions in the image throughout the forward pass.




In summary, the workflow of adaptive downsampling is as follows: 1) obtain importance score $s_i=\sigma(l(f_i))$ for each token $f_i$, 2) calculate the grid prior $g_i$ for each token, 3) pick the top-$x$\% (e.g., 1/4 or 1/5) tokens with highest $g_i+\alpha  s_i$ values;
4) perform neighborhood merging around the location of the top-$x$\% tokens using the formulation given in Eq.~\eqref{eq:merging} and obtain the merged tokens for the next stage.

\section{Implementation Details}
\subsection{Point-based versions of segmentation heads}\label{sec:mask2former}

Traditional segmentation heads only operate on a set of rectangular hierarchical feature maps. In our case, the output of the backbone is a set of features on 
irregularly spaced tokens. To compare with prior work, we convert two recent Transformer-based segmentation heads, Mask2Former and HCFormer, so as to operate on irregularly spaced tokens. 

Mask2Former~\cite{mask2} uses the multi-scale deformable attention from deformable DETR~\cite{deformable} as part of its pixel decoder. In the multi-scale deformable attention layer, a token in one feature map attends to $k$ locations in every hierarchical level. It learns the offset $\Delta p_i$ between its own location $p_c$ and the $i$-th sampling location $p_i=p_c+\Delta p_i$ by performing bilinear interpolation on the $2\times 2$ patch surrounding $p_i$. We replace the bilinear interpolation with an inverse distance weighting-based interpolation, with a learnable power initialized at 6. Specifically, we gather 4 tokens closest in Euclidean space to $p_c+\Delta p_i$, and use the inverse of their distances to $p_c+\Delta p_i$ raised to the learned power for a weighted average of their features (weights sum normalized to 1), to obtain the feature of the sampled neighbor. In addition, we replace the $3\times 3$ convolutions in the pixel decoder by a PointConv layer. We implement the weight net $w(\cdot)$ in all the PointConv layers used in our model (including those in adaptive downsampling) as one fully-connected layer followed by one LayerNorm and one GELU activation. The mid-channel number $C_{\text{mid}}$ is set to 4.

We introduce the other segmentation head HCFormer~\cite{hcformer} in the supplementary material.

\subsection{Scale and Rotation Invariance}

Our adaptive downsampling mechanism allows us to utilize the same number of features for both small and large scale objects. 
However, this creates a difficulty for the position embedding to achieve scale invariance, since tokens on a smaller object are closer together and have smaller relative distances, resulting in vastly different position embedding values than tokens on a large object. 

To address this concern, we expand the relative position vector from $(\Delta x, \Delta y) = (x_{i} - x_{j}, y_{i} - y_{j})$ to
\begin{equation}
    \!\left(\!\Delta x, \Delta y, \sqrt{\Delta x^2 \!+\! \Delta y^2}, \dfrac{\Delta x}{\sqrt{\Delta x^2 \!+\! \Delta y^2}}, \dfrac{\Delta y}{\sqrt{\Delta x^2 \!+\! \Delta y^2}}\!\right)\!.
\end{equation}
Note, the latter three terms are distance, cosine and sine values of the relative position. Distance is rotation-invariant and the angle values are scale-invariant. These expanded terms facilitate learning appropriate scale-invariant and rotational-invariant embeddings.

\subsection{Blank Tokens}

During early experiments on ImageNet, we observed abnormally large feature norms for tokens in texture-less corners of the images, usually far away from any  object, both for our model and for Swin~\cite{swin} Transformers. We suspect this is because of the strong gradient softmax has when it cannot separate near-identical (yet irrelevant) features. 
To eliminate this artifact, we introduced a learnable \textit{blank token} $(K_{\text{blank}}^{ij},V_{\text{blank}}^{ij})\in \mathbb{R}^{C}\times \mathbb{R}^{C}$ shared by all neighborhoods in the $j$-th transformer block in the $i$-th stage. Thus, when there is no useful content in the neighborhood, the softmax operator can learn to simply attend to the blank token and avoid attempting to distribute attention in textureless regions.  



\begin{table*}
\begin{center}
\begin{small}
\begin{tabular}{lcccccccc}\hline
Variant & \# Blocks & Dim & Heads & \makecell{MLP\\Ratio} & \makecell{Cluster\\ Size} & \makecell{Neighborhood\\Size} & \# Params & FLOPs\\\hline
AFF-Mini & 2,2,6,2 & 32,128,256,384 & 2,4,8,16 & 2 & 8 & 48 & 6.75M & 1.08G \\
AFF-Tiny & 3,4,18,5 & 64,128,256,512 & 2,4,8,16 & 3 & 8 & 48 & 27.02M & 4.03G \\
AFF-Small & 3,4,18,2 & 96,192,384,768 & 3,6,12,24 & 3 & 8 & 48 & 42.61M & 8.16G \\
AFF-Base & 3,4,18,2 & 128,256,512,1024 & 4,8,16,32 & 3 & 24 & 144 & 75.34M & 42.54G \\\hline 
\end{tabular}
\end{small}
\end{center}
\vspace{-0.6cm}
\caption{Configurations of AFF.}
\label{tb:config}
\vskip -0.15in
\end{table*}

\begin{table}
\begin{center}
\begin{footnotesize}
\begin{tabular}{c|l@{ }c@{ }c@{ }c}
& \multicolumn{1}{l}{Model} & Top-1 Acc & \# Params & FLOPs\\\hline
\multirow{7}{*}{\rotatebox[origin=c]{90}{Mini}} & A-ViT-T+distl.~\cite{avit} & 72.4\% & 5M & 0.8G \\
&\makecell[l]{Token Pooling DeiT-e318
\\Sparsity level 5~\cite{tokenpooling}} & 76.8\% & - & 1.1G\\
&Swin-Mini$^{\ddagger}$ & 76.9\% & 6.76M & 1.07G \\
&EdgeViT-XS~\cite{edgevit} & 77.5\% & 6.7M & 1.1G \\
\cdashline{2-5} 
&AFF-Mini & \textbf{78.2\%} & 6.75M & 1.08G \\
&AFF-Mini-1/5 & 77.5\% & 6.75M & \textbf{0.72G} \\
\hline
\multirow{12}{*}{\rotatebox[origin=c]{90}{Tiny}}&DynamicViT-DeiT-S/0.7~\cite{dynamicvit} & 79.3\% & - & 2.9G \\
&TokenLearner S/32(22)~\cite{tokenlearner} & 79.4\% & - &3.3G\\
&A-ViT-S+distl.~\cite{avit} & 80.7\% & 22M & 3.6G \\
&AdaViT~\cite{adavit} & 81.1\% & - & 3.9G \\
&PS-ViT-B/14~\cite{psvit} & 81.7\% & 21.3M & 5.4G \\
&LightViT-B~\cite{lightvit} & 82.1\% & 35.2M & 3.9G \\
&PVT v2-B2~\cite{pvtv2} & 82\% & 25.4M & 4G \\
&Swin-Tiny & 81.3\% & 28M & 4.5G \\
&Swin-Tiny$^{\ddagger}$ & 81.9\% & 27M & 4G \\
&ConvNeXt-Tiny~\cite{convnext} & 82.1\% & 28M & 4.5G \\
\cdashline{2-5} 
&AFF-Tiny & \textbf{83\%} & 27M & 4G \\
&AFF-Tiny-1/5 & 82.4\% & 27M & \textbf{2.74G} \\
\hline 
\multirow{7}{*}{\rotatebox[origin=c]{90}{Small}}&Swin-Small & 83\% & 50M & 8.7G \\
&Swin-Small$^{\ddagger}$ & 82.9\% & 42.6M & 8.14G \\
&ConvNeXt-Small~\cite{convnext} & 83.1\% & 50M & 8.7G \\
&PS-ViT-B/18~\cite{psvit} & 82.3\% & 21.3M & 8.8G \\
&TokenLearner B/32(20)~\cite{tokenlearner} & 82.7\% & - & 11.5G \\
\cdashline{2-5} 
&AFF-Small & \textbf{83.5\%} & 42.6M & 8.16G \\
&AFF-Small-1/5 & \textbf{83.4\%} & 42.6M & \textbf{5.69G} \\
\hline 
\multirow{2}{*}{\rotatebox[origin=c]{90}{Base}}&Swin-Base$^{\dagger}$ & 86.4\% & 88M & 47.0G \\
&AFF-Base$^{\dagger}$ & 86.2\% & 75.3M & 42.5G \\
\hline
\end{tabular}
\end{footnotesize}
\end{center}
\vspace{-0.6cm}
\caption{ImageNet Top-1 validation accuracy comparison at $224\times 224$ resolution. ``1/5" means the model uses 1/5 downsampling rate instead of the traditional 1/4 downsampling rate. ``-" means not reported. 
The Swin backbones$^\ddagger$ are trained using the same architecture configuration and training settings as our model.
The base models$^{\dagger}$ are pre-trained on ImageNet-22K and subsequently fine-tuned on ImageNet-1K at $384\times 384$ resolution. }
\label{tb:imagenet}
\vskip -0.1in
\end{table}

\section{Experiments}\label{sec:expe}

\subsection{Image Classification on ImageNet-1K}
We evaluate on image classification using ImageNet-1K~\cite{imagenet}, which contains 1.28M training images and 50K validation images from 1000 classes. 

\noindent\textbf{Settings.} Table~\ref{tb:config} contains specifications for the different model sizes used in our comparisons. Unless stated otherwise, we use the ``1/4'' downsampling rate, cluster size 8 and neighborhood size 48 (i.e., we collect neighbors from the nearest 6 clusters for each token). For ImageNet only, we switch to global attention in the last stage, because at an input resolution of  $224\times 224$, the last stage only has 49 tokens left. We set the hyperparameter $\alpha=4$ in the merging center score calculation.
We largely follow the training hyperparameters from Swin Transformers. We train for 300 epochs. 

\noindent\textbf{Results.} We present results in Table~\ref{tb:imagenet}, divided into sections according to model sizes. 
Compared to Swin~\cite{swin}, we obtain +1.3\%, +1.7\% and +0.5\% improvement for Mini, Tiny and Small, respectively. For a fair comparison, we trained Swin-Tiny and Swin-Small using our own architecture configuration, training settings and patch embedding layer, and we still obtain +1.1\% and +0.6\% improvement, respectively. The difference is not large, which is expected because adaptive downsampling is mainly geared toward dense prediction tasks. We did outperform all previous adapative downsampling approaches with global attention such as AdaViT, DynamicViT and A-ViT.

We further demonstrate our model's capability to use flexible downsampling rates by showing the results with a 1/5 downsampling rate. Across all sizes, the 1/5 downsampling rate brings more than 30\% drop in FLOP count compared to the 1/4 downsampling counterparts, with minimal accuracy drop and still beating all Swin baselines.

\subsection{Segmentation}
\noindent\textbf{Datasets.} We evaluate on semantic, instance, and panoptic segmentation using 3 datasets: ADE-20K~\cite{ade} is a semantic segmentation dataset containing 150 categories across 20K training images and 2K validation images. Cityscapes~\cite{cityscapes} is a street-view dataset with high quality annotations, containing 2975 training images and 500 validation images, with a total of 19 object classes. COCO 2017~\cite{coco} is an instance segmentation dataset, containing 118K training and 5K validation images.

\noindent\textbf{Settings.} We mostly follow the settings of Mask2Former~\cite{mask2}. Please see the supplementary for details.
Tasks with Swin-Mini backbone are trained under the same settings as AFF-Mini. Other Swin Transformer results are taken from \cite{mask2}.

\noindent\textbf{Results.} We present the results of semantic segmentation on ADE20K~\cite{ade} (Table~\ref{tb:ade}), instance and panoptic segmentation on Cityscapes~\cite{cityscapes} (Table~\ref{tb:cityscapes}) and instance segmentation on  COCO~\cite{coco} (in supplementary).

For semantic segmentation on ADE20K (Table~\ref{tb:ade}), we show mIoU improvement across all three sizes compared to the Swin baselines: +2.4\% for Mini, +1.9\% for Mini-1/5; +2.5\% for Tiny, +2.3\% for Tiny-1/5; and +0.6\% for Small-1/5. Our 1/5 downsampling-rate models decrease FLOP count by 20\%, while improving over Swin across model sizes. We furthermore isolate the effect of changing the Mask2Former head to operate on point clouds (Section \ref{sec:mask2former}). Using Swin-Mini on semantic segmentation, we find the modified head reduces FLOP count (-9.4\%) with slightly lower  performance (-0.4\%).


For instance and panoptic segmentation on Cityscapes (Table~\ref{tb:cityscapes}), both our Tiny and Small models demonstrate a solid improvement over the baselines with the Swin backbone. For Tiny, our model improves the Panoptic PQ metric by +1.8\%, and the Instance AP by +3.0\%. For Small, our model improves the Panoptic PQ metric by +2.1\%, and the Instance AP by +2.2\%. Note that the overall performance of AFF-Tiny is \textbf{on par with Swin-Base, a model 3.3x larger}; AFF-Small is \textbf{on par with Swin-Large, a model 4.6x larger}, while both baselines are pre-trained with $\mathbf{10}$ times more data on  ImageNet22k~\cite{imagenet21k}.

\begin{table}
\begin{center}
\begin{footnotesize}
\begin{tabular}{llccc}
Backbone & Segmentation Head & \makecell{Crop\\Size} & mIoU & FLOPs\\\hline
Swin-Mini$^{\ddagger}$ & Mask2Former~\cite{mask2} & 512 & 44.5 & 54G \\
Swin-Mini$^{\ddagger}$ & Mask2Former*~\cite{mask2} & 512 & 44.1 & 48.9G \\
\hdashline
AFF-Mini & Mask2Former*~\cite{mask2} & 512 & \textbf{46.5} & 48.3G \\
AFF-Mini-1/5 & Mask2Former*~\cite{mask2} & 512 & \textbf{46.0} & \textbf{39.9G} \\
\hline
MiT-B2 & SegFormer~\cite{segformer} & 512 & 46.5 & 62.4G \\
PVT v2-B3~\cite{pvtv2} & Semantic FPN~\cite{semanticfpn} & 512 & 47.3 & 62.4G \\
Swin-Tiny & Mask2Former~\cite{mask2} & 512 & 47.7 & 74G \\
\hdashline
AFF-Tiny & Mask2Former*~\cite{mask2} & 512 &  \textbf{50.2} & 64.6G \\
AFF-Tiny-1/5 & Mask2Former*~\cite{mask2} & 512 & \textbf{50.0} & \textbf{51.1G} \\
\hline
PVT v2-B5~\cite{pvtv2} & Semantic FPN~\cite{semanticfpn} & 512 & 48.7 & 91.9G \\
MiT-B5 & SegFormer~\cite{segformer} & 640 & 51 & 183.3G \\
Swin-Small & Mask2Former~\cite{mask2} & 512 & 51.3 & 98G \\
\hdashline
AFF-Small & Mask2Former*~\cite{mask2} & 512 & 51.2 & 87G \\
AFF-Small-1/5 & Mask2Former*~\cite{mask2} & 512 & \textbf{51.9} & \textbf{67.2G} \\
\hline
\end{tabular}
\end{footnotesize}
\end{center}
\vspace{-0.6cm}
\caption{Semantic segmentation on ADE20K val. ``1/5": backbone uses 1/5 downsampling rate instead of 1/4. *~: Segmentation head modified to accept point cloud input. 
The Swin backbone$^\ddagger$ is trained using the same architecture configuration and training settings as our model. Seed fixed at 0.
}
\label{tb:ade}
\vskip -0.1in
\end{table}


\begin{table}
\begin{center}
\begin{footnotesize}
\begin{tabular}{llcccc}
Backbone & \makecell{Segmentation\\Head} & \makecell{Panoptic \\PQ (s.s.)} & \makecell{Instance\\AP} & \makecell{Backbone\\\# Params} 
\\\hline
AFF-Mini & Mask2Former*~\cite{mask2} & 62.7 & 40.0 & 6.75M 
\\\hdashline
Swin-Tiny & Mask2Former~\cite{mask2} & 63.9 & 39.7 & 28M 
\\
AFF-Tiny & Mask2Former*~\cite{mask2} & \textbf{65.7} & \textbf{42.7} & 27M 
\\
\hdashline
Swin-Small & Mask2Former~\cite{mask2} & 64.8 & 41.8 & 50M\\ 
AFF-Small & Mask2Former*~\cite{mask2} & \textbf{66.9} & \textbf{44.0} & 42.6M\\ \hdashline
Swin-Base$^\dagger$ & Mask2Former~\cite{mask2} & 66.1 & 42 & 88M \\
AFF-Base$^\dagger$ & Mask2Former*~\cite{mask2} & \textbf{67.7} & \textbf{46.2} & 75.3M\\ \hdashline
Swin-Large$^\dagger$ & Mask2Former~\cite{mask2} & 66.6 & 43.7 & 197M\\ 
\hline
\end{tabular}
\end{footnotesize}
\end{center}
\vspace{-0.55cm}
\caption{Segmentation on Cityscapes. We set $\alpha=8$. * The segmentation head is modified to accept point cloud input. $^\dagger$ Backbone pre-trained with ImageNet-22K. Seed fixed at 0.}
\label{tb:cityscapes}
\vskip -0.15in
\end{table}

\subsection{Ablations}

In Table~\ref{tb:ablation}, we ablate design choices using AFF-Mini. For example, replacing our adaptive downsampling with the downsampling approach in Swin\cite{swin} reduces top-1 ImageNet accuracy by 0.5\%. Replacing our expanded position embedding with a baseline position embedding -- passing only $(x,y)$ differences as input to the position embedding function -- reduces top-1 ImageNet accuracy by 0.9\%. 

\begin{table}
\begin{center}
\begin{footnotesize}
\begin{tabular}{lcc}
Variant & \makecell{Downsampling\\rate} & \makecell{ImageNet1K\\Top-1 Acc}
\\\hline
Full model & 1/5 & \textbf{77.5\%}\\
Remove grid prior & 1/5 & 76.6\%\\
Remove expanded relative position & 1/5 & 76.6\%\\
Remove blank token & 1/5 & 76.9\%\\
Remove reserved tokens & 1/5 & 77.2\%\\
\hline
Full model & 1/4 & \textbf{78.2\%}\\
Use PatchMerging from Swin~\cite{swin} & 1/4 & 77.7\%\\
\hline 
\end{tabular}
\end{footnotesize}
\end{center}
\vspace{-0.6cm}
\caption{Ablation studies based on AFF-Mini. 
}
\vskip -0.2in
\label{tb:ablation}
\end{table}

In Table~\ref{tb:alpha}, we present how the hyperparameter $\alpha$ influences the Top-1 accuracy on ImageNet and instance segmentation AP on Cityscapes~\cite{cityscapes}. 
$\alpha$ is used to balance the weight between the grid prior $g_i$ and the importance score $s_i$. The larger the $\alpha$ is, the more ``adaptive" the remaining tokens are. 
For ImageNet classification, a larger $\alpha$ results in minor degradation in performance, while a reasonably large $\alpha$ results in better performance for instance segmentation. This is expected, as our adaptive downsampling module allocates more tokens to smaller objects and details around object boundaries. While this might not be needed by the usually centered objects in ImageNet, it is particularly beneficial for instance segmentation. 

\begin{table}[htb]
\begin{center}
\begin{footnotesize}
\begin{tabular}{c|cc}
\makecell{$\alpha$} & ImageNet-1K Acc &\makecell{Cityscapes\\Instance AP}
\\\hline
2 & 78.4\% & 37.5 \\
4 & 78.2\% & 38.7 \\
6 & - & 38.7 \\
8 & - & \textbf{40.0} \\
10 & - & 37.8 \\
12 & - & 38.1 \\
\hline
\end{tabular}
\end{footnotesize}
\end{center}
\vspace{-0.6cm}
\caption{Ablation studies on $\alpha$ using AFF-Mini and Mask2Former~\cite{mask2} heads. Larger $\alpha$ means larger weight on the learned importance score $s_i$,  compared to grid prior $g_i$. For segmentation tasks with $\alpha$ larger than 4, we use the $\alpha=4$ ImageNet pretrained checkpoint and modify $\alpha$ during fine-tuning. }
\label{tb:alpha}
\vskip -0.15in
\end{table}


\section{Conclusion}\label{sec:conclusion}
In this paper we propose AutoFocusFormer, a novel image recognition backbone that is to our knowledge the first local attention transformer with successive adapative downsampling stages for segmentation tasks. We proposed to perform local attention on neighborhoods generated from  balanced clusters obtained with a novel approach based on space-filling curves, as well as a novel learnable adaptive downsampling algorithm that automatically locates the important regions of the image. Besides, we adapted state-of-the-art segmentation heads to be able to utilize a set of irregular tokens. Experiments show that our algorithm improves significantly over baselines for all segmentation tasks, while offering more flexible downsampling rates during training and inference time. We believe our backbone could inspire models for other tasks that would want to focus on important image locations and benefit from the non-grid structure.
\vspace{-0.25in}
\subsubsection*{Acknowledgements}
We thank Dr. Hanlin Goh and Dr. Tatiana Likhomanenko for valuable suggestions to improve the paper. Chen Ziwen is partially supported by NSF grant $\#1751402$.

{\small
\bibliographystyle{ieee_fullname}
\bibliography{egbib}
}

\clearpage

\appendix

\noindent                                                                                                               \textbf{\huge Appendix}
\vspace{0.25in}

\section{Additional Experimental Results}
\subsection{Additional Ablation Studies}

\subsubsection{Ablation on Balanced Clustering Algorithm}

We show more results and comparisons regarding our balanced clustering algorithm in Table~\ref{tb:cluster}. We study the benefits of space-filling anchors and different types of space-filling curves. We measure the quality of the resulting clusters using the silhouette coefficient~\cite{silhouettes} metric. The silhouette coefficient ranges from $-1$ to $1$, measuring  how clearly distinguishable the clusters are. A larger value indicates better clusters. Specifically, the silhouette score for the $i$-th token is calculated as
\begin{equation}
    \dfrac{b_i-a_i}{\max(a_i,b_i)},
\end{equation}
where $a_i$ is the mean distance between the position of the $i$-th token and all other tokens in the same cluster, and $b_i$ is the mean distance between the position of the $i$-th token and all tokens in the next nearest cluster. The final silhouette coefficient is the average score of all the tokens. The numbers in Table~\ref{tb:cluster} are averaged over a random batch of 256 images from ImageNet.

Our default setting is to use space-filling anchors, and apply a simple horizontal scanline as the space-filling curve on the anchors. A horizontal scanline sweeps the rows from left to right in odd rows and from right to left in even rows. We experimented with two other more complicated curves here: the Peano~\cite{peano} and the Hilbert~\cite{hilbert} curves. Both are recursive curves establishing a surjective mapping from a unit interval to a unit square. We also studied direct application of space-filling curves to tokens without us of the anchors. Results show that the anchors are necessary for obtaining more  separated clusters. Also surprisingly, the simple horizontal scanline attains better cluster quality than the more complicated space-filling curves when the anchors are used. The visualization of the clustering results are shown in Fig.~\ref{fig:cluster}.

\begin{table}
\begin{center}
\begin{footnotesize}
\begin{tabular}{c@{ }c|c@{ }c@{ }c@{ }c}
& & \multicolumn{4}{c}{Silhouette Coefficient $\uparrow$}\\\cline{3-6}
Anchors & \makecell{Space-filling\\curve type} &  \makecell{Stage 1} & \makecell{Stage 2} & \makecell{Stage 3} & \makecell{Stage 4}
\\\hline
\cmark & \makecell{horizontal\\scanline} & \textbf{0.24} & \textbf{0.24} & 0.22 & \textbf{0.24}\\
\xmark & \makecell{horizontal\\scanline} & -0.01 & -0.20 & -0.16 & 0.03\\
\cmark & Peano & 0.2 & 0.23 & \textbf{0.29} & 0.21 \\
\xmark & Peano & 0.15 & 0.15 & 0.14 & 0.17 \\
\cmark & Hilbert & 0.22 & 0.23 & 0.22 & 0.19 \\
\xmark & Hilbert & 0.14 & 0.15 & 0.16 & 0.18 \\
\hline
\end{tabular}
\end{footnotesize}
\end{center}
\vspace{-0.6cm}
\caption{Ablation studies on the anchors and the type of space-filling curves used in the balanced clustering algorithm. For the cases without anchors, the space-filling curve is directly applied on the tokens. The metric scores are averaged over a random batch of 256 images from ImageNet.}
\label{tb:cluster}
\end{table}

\begin{table*}
\begin{center}
\begin{footnotesize}
\begin{tabular}{@{}l@{\hspace{2mm}}l@{\hspace{2mm}}c@{\hspace{2mm}}c@{\hspace{2mm}}c@{\hspace{2mm}}c@{\hspace{2mm}}c@{\hspace{2mm}}c@{\hspace{2mm}}c@{\hspace{2mm}}c@{\hspace{2mm}}}
Backbone & Segmentation Head & Search Space & Epochs & AP & AP\textsuperscript{S} & AP\textsuperscript{M} & AP\textsuperscript{L} 
& \# Params & FLOPs\\\hline
EdgeViT-XS~\cite{edgevit} & Mask R-CNN~\cite{maskrcnn} & - & 12 & 38.3 & - & - & - & 26.5M & - \\
PVT v2-B1~\cite{pvtv2} & Mask R-CNN~\cite{maskrcnn} & - & 12 & 38.8 & - & - & - & 33.7M & - \\
LightViT-T~\cite{lightvit} & Mask R-CNN~\cite{maskrcnn} & - & 36 & 38.4 & - & - & - & 28M & 187G \\
Swin-Mini$^{\ddagger}$ & Mask2Former*~\cite{mask2} & 100 queries & 50 & 33.1 & 13.8 & 35.2 & 53.7 & 25.8M & 149G \\
\hdashline
AFF-Mini & Mask2Former*~\cite{mask2} & 100 queries & 50 & \textbf{42.3} & 21.2 & \textbf{45.6} & 63.7 & 25.8M & 148G \\
AFF-Mini-1/5 & Mask2Former*~\cite{mask2} & 100 queries & 50 & \textbf{42.3} & \textbf{21.8} & \textbf{45.7} & \textbf{64.0} & 25.8M & \textbf{120G (-19\% vs. Swin)} \\
\hline
PVT v2-B3~\cite{pvtv2} & Mask R-CNN~\cite{maskrcnn} & - & 12 & 42.5 & - & - & - & 64.9M & - \\
LightViT-S~\cite{lightvit} & Mask R-CNN~\cite{maskrcnn} & - & 36 & 39.9 & - & - & - & 38M & 204G \\
SpineNet-96~\cite{spinenet} & Mask R-CNN~\cite{maskrcnn} & 1000 proposals & 350 & 41.5 & - & - & - & 55.2M & 315G \\
Swin-Tiny & Mask2Former~\cite{mask2} & 100 queries & 50 & 45.0 & 24.5 & 48.3 & \textbf{67.4} & 47M & 232G \\
\hdashline
AFF-Tiny & Mask2Former*~\cite{mask2} & 100 queries & 50 & \textbf{45.3} & \textbf{24.8} & \textbf{49.2} & 66.9 & 46M & 204G (-12\% vs. Swin) \\
AFF-Tiny-1/5 & Mask2Former*~\cite{mask2} & 100 queries & 50 & 44.5 & 24.5 & 47.8 & 66.3 & 46M & \textbf{152G (-34\% vs. Swin)}\\
\hline
LightViT-B~\cite{lightvit} & Mask R-CNN~\cite{maskrcnn} & - & 36 & 41.2 & - & - & - & 54M & 240G \\
PVT v2-B5~\cite{pvtv2} & Mask R-CNN~\cite{maskrcnn} & - & 12 & 42.5 & - & - & - & 101.6M & - \\
SpineNet-190~\cite{spinenet} & Mask R-CNN~\cite{maskrcnn} & 1000 proposals & 500 & 46.1 & - & - & - & 176.2M & 2077G \\
Swin-Small & Mask2Former~\cite{mask2} & 100 queries & 50 & \textbf{46.3} & 25.3 & \textbf{50.3} & \textbf{68.4} & 69M & 313G \\
\hdashline
AFF-Small & Mask2Former*~\cite{mask2} & 100 queries & 50 & \textbf{46.4} & \textbf{27.0} & 49.8 & 67.6 & 61.4M & 281G (-10\% vs. Swin) \\
AFF-Small-1/5 & Mask2Former*~\cite{mask2} & 100 queries & 50 & 45.7 & 26.1 & 49.2 & 67.5 & 61.4M & \textbf{206G (-34\% vs. Swin)} \\
\hline
\end{tabular}
\end{footnotesize}
\end{center}
\vspace{-0.6cm}
\caption{Instance segmentation on COCO instance val2017. ``1/5" means the backbone uses 1/5 downsampling rate instead of the traditional 1/4 downsampling rate. * The segmentation head is modified to accept point cloud input. 
$^\ddagger$~This Swin backbone is trained using the same architecture configuration and training settings as our model. 
The random seed is fixed at 0. 
}
\label{tb:instance}
\vskip -0.15in
\end{table*}

\begin{table}
\begin{center}
\begin{footnotesize}
\begin{tabular}{lcccc}
Backbone & Segmentation Head & \makecell{Crop\\Size} & mIoU & FLOPs\\\hline
Swin-Small & HCFormer~\cite{hcformer} & 512 & 48.8 & 56G \\
PVT v2-B5~\cite{pvtv2} & Semantic FPN~\cite{semanticfpn} & 512 & 48.7 & 91.9G \\
\hdashline
AFF-Small & HCFormer*~\cite{hcformer} & 512 & \textbf{49.2} & \textbf{51.1G} \\
\hline
\end{tabular}
\end{footnotesize}
\end{center}
\vspace{-0.6cm}
\caption{Semantic segmentation on ADE20K val with HCFormer head. *~The segmentation head is modified to accept point cloud input. }
\label{tb:ade_hcformer}
\vskip -0.15in
\end{table}

\subsection{Additional Segmentation Experiments}

\subsubsection{Additional Results on COCO and ADE20K}


\textbf{COCO instance segmentation.} For instance segmentation on COCO (Table~\ref{tb:instance}), we present very significant AP improvement for  the Mini size, showing the capability of our model of being more efficient with limited resources. For Tiny and Small, we obtained par results with Swin with 10\% decrease in FLOPs. For the 1/5 downsampling-rate models, we see they have significant computational benefits with little performance drop with respect to their 1/4 counterparts. 
We observe AP improvements for small objects (AP\textsuperscript{S}) and regressions for large objects (AP\textsuperscript{L}). We suspect that the standard decoder heads were not aggregating information well when the sampling rate is very uneven for large objects, and we aim to improve the decoder in future work.

\textbf{Semantic segmentation with HCFormer.} HCFormer~\cite{hcformer} performs prediction on the feature map at the coarsest level. Each token on the finer level will learn 9 similarity values to the tokens in a $3\times 3$ window in the coraser level, and the model uses the similarity values to interpolate the prediction all the way up to the highest resolution. We replace the square window by 9 nearest neighbors in the coarser level, while the calculation of similarity values stays the same. 
In Table~\ref{tb:ade_hcformer}, we show semantic segmentation results on the ADE20K~\cite{ade} dataset with the HCFormer~\cite{hcformer} head for the AFF-Small model. We achieve a +0.4\% increase in the mIoU metric with -8\% FLOP count.

\begin{table}
\begin{center}
\begin{footnotesize}
\begin{tabular}{l||c|c||c|c}
Class & Swin-Tiny & AFF-Tiny & Swin-Small & AFF-Small\\\hline
person  & 36.3 & 38.6 & 36.8 & 39.2 \\
rider &  29.0  & 30.4 & 28.7 & 33.3\\
car  &   59.3 & 60.9 & 59.9 & 61.4\\
truck &  41.4 & 43.1 & 42.0 & 41.1\\
bus &  60.4   & 65.1 & 65.2 & 66.9\\
train &  43.7 & 52.3 & 51.7 & 55.4\\
motorcycle & 24.5 & 26.2 & 25.2 & 28.9\\
bicycle &  23.2   & 24.9 & 24.7 & 25.6\\\hline
average &   39.7 & 42.7 & 41.8 & 44.0 \\
\end{tabular}
\end{footnotesize}
\end{center}
\vspace{-0.6cm}
\caption{Class-wise Instance Segmentation AP on CityScapes (backbone Swin vs.\ AFF) with Mask2Former  segmentation head.}
\label{tb:csclass}
\vskip -1em
\end{table}

\begin{table*}
\begin{center}
\begin{footnotesize}
\begin{tabular}{l|c|c||l|c|c||l|c|c}
Class & Swin-Tiny & AFF-Tiny & Class & Swin-Tiny & AFF-Tiny & Class & Swin-Tiny & AFF-Tiny \\\hline
person        & 50.541 & 50.210 & bicycle      & 23.690 & 23.861 & car            & 45.443 & 46.281 \\
 motorcycle    & 40.718 & 40.840 & airplane     & 60.491 & 59.508 & bus            & 70.112 & 71.876 \\
 train         & 71.873 & 71.853 & truck        & 42.820 & 44.763 & boat           & 29.597 & 30.855 \\
 traffic light & 30.971 & 30.314 & fire hydrant & 70.358 & 68.929 & stop sign      & 68.312 & 68.553 \\
 parking meter & 50.348 & 50.078 & bench        & 24.617 & 25.122 & bird           & 33.864 & 34.639 \\
 cat           & 76.870 & 77.594 & dog          & 68.567 & 70.120 & horse          & 47.912 & 47.360 \\
 sheep         & 53.354 & 54.451 & cow          & 56.130 & 55.105 & elephant       & 66.213 & 65.602 \\
 bear          & 77.146 & 81.873 & zebra        & 65.191 & 66.634 & giraffe        & 61.708 & 61.146 \\
 backpack      & 23.438 & 23.936 & umbrella     & 54.531 & 53.804 & handbag        & 22.801 & 24.080 \\
 tie           & 37.348 & 36.612 & suitcase     & 50.478 & 50.670 & frisbee        & 68.770 & 69.198 \\
 skis          & 7.103  & 7.668  & snowboard    & 31.047 & 31.945 & sports ball    & 50.537 & 50.470 \\
 kite          & 38.308 & 38.900 & baseball bat & 38.899 & 38.439 & baseball glove & 45.918 & 48.275 \\
 skateboard    & 37.358 & 41.363 & surfboard    & 40.484 & 41.238 & tennis racket  & 61.187 & 61.463 \\
 bottle        & 42.015 & 42.877 & wine glass   & 37.538 & 38.286 & cup            & 47.234 & 49.267 \\
 fork          & 23.725 & 24.974 & knife        & 18.923 & 19.837 & spoon          & 19.997 & 22.391 \\
 bowl          & 45.014 & 45.610 & banana       & 27.437 & 25.364 & apple          & 24.022 & 25.128 \\
 sandwich      & 47.375 & 47.423 & orange       & 37.039 & 37.065 & broccoli       & 24.394 & 25.494 \\
 carrot        & 24.957 & 24.435 & hot dog      & 45.251 & 41.675 & pizza          & 58.760 & 57.535 \\
 donut         & 55.920 & 57.046 & cake         & 49.705 & 48.989 & chair          & 26.461 & 27.313 \\
 couch         & 48.393 & 47.571 & potted plant & 27.185 & 26.932 & bed            & 45.991 & 44.751 \\
 dining table  & 22.249 & 23.094 & toilet       & 68.757 & 68.878 & tv             & 67.036 & 66.700 \\
 laptop        & 69.644 & 70.101 & mouse        & 66.043 & 61.897 & remote         & 39.732 & 40.769 \\
 keyboard      & 55.528 & 56.306 & cell phone   & 41.274 & 42.047 & microwave      & 65.497 & 64.513 \\
 oven          & 39.090 & 38.031 & toaster      & 40.422 & 39.761 & sink           & 41.816 & 42.876 \\
 refrigerator  & 66.878 & 66.622 & book         & 15.363 & 16.021 & clock          & 56.141 & 57.378 \\
 vase          & 41.719 & 42.520 & scissors     & 35.620 & 37.310 & teddy bear     & 55.229 & 52.801 \\
 hair drier    & 9.155  & 14.401 & toothbrush   & 28.801 & 26.657 &                &        &        \\ 
\end{tabular}
\end{footnotesize}
\end{center}
\vspace{-0.6cm}
\caption{Class-wise Instance Segmentation AP on COCO (backbone Swin vs.\ AFF) with Mask2Former  segmentation head.}
\label{tb:cococlass}
\vskip -1em
\end{table*}

\subsubsection{Class-wise Segmentation Results}

To facilitate understanding how AFF improves over the baselines, in addition to score breakdown according to object sizes, we further provide class-wise segmentation score breakdown in Table~\ref{tb:csclass} and Table~\ref{tb:cococlass}. However, through theses results, we don't see apparent correlation between score improvement and classes. We guess that the improvement from AFF is more correlated with object sizes than specific categories.

\section{Segmentation Training Setting Details}\label{sec:setting}
We largely follow the settings of Mask2Former~\cite{mask2} in training including weight decay, augmentations and training steps. 
More specifically, we use the AdamW~\cite{adamw} optimizer with the step learning rate scheduler. We use a weight decay of 0.05. We apply a learning rate multiplier 0.1 to the backbone. We set $\alpha=4$ for ADE20K and COCO, and $\alpha=8$ for Cityscapes. We use a learnable shepard power initialized at 6 for ADE20K and Cityscapes, and a fixed power 4 for COCO.

For ADE20K, we train for 80K steps with a batch size of 32 and a base learning rate 0.0002. The FLOP count is calculated on a random $512\times 512$ image, as we crop all images to this size during training.

For COCO, we train for 50 epochs with a batch size of 64 and a base learning rate 0.0002. We apply the large-scale jittering (LSJ) augmentation~\cite{du2021simple, simplycopy} with a random scale sampled from range 0.1 to 2.0 followed by a fixed size crop to $1024\times 1024$ during training. During inference we use the standard Mask R-CNN~\cite{maskrcnn} inference setting where we resize an image with shorter side to 800 and longer side up to 1333. The FLOP count is averaged over 100 validation images for the COCO FLOP count. We scale the learning rate down by 0.1 at 0.9 and 0.95 fractions of the total training steps.

For Cityscapes, we train for 45K steps with a batch size of 32 and a base learning rate 0.0002. During training, we use a crop size of $512 \times 1024$. During inference, we use the entire image ($1024 \times 2048$). We use 100 queries for all models.

For all training tasks, we do not use test-time augmentation or multi-scale testing. For all segmentation results, we report the best validation result in one run with seed fixed at 0. Validation results are reported every 2500 steps.

\section{Qualitative Comparisons}

In Fig.~\ref{fig:city}, we provide a qualitative comparison of AFF-Small and Swin-Small with Mask2Former segmentation head on the Cityscapes panoptic segmentation data, along with the remaining token locations in stage 2, 3 and 4. Our model is able to retain tokens on very small objects even in the last stage, which provides the foundation to capture crowded, small objects, such as the people sitting in the cafe in the first example in Fig.~\ref{fig:city}.

\begin{figure*}
    \centering
    \begin{subfigure}{0.32\linewidth}
        \includegraphics[width=1.0\columnwidth]{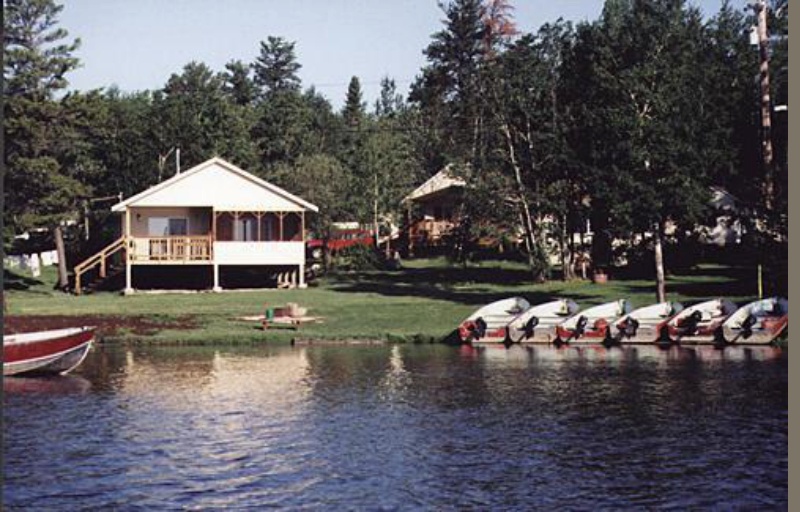} 
   \end{subfigure}
   \hfill
   \begin{subfigure}{0.32\linewidth}
    \includegraphics[width=1.0\columnwidth]{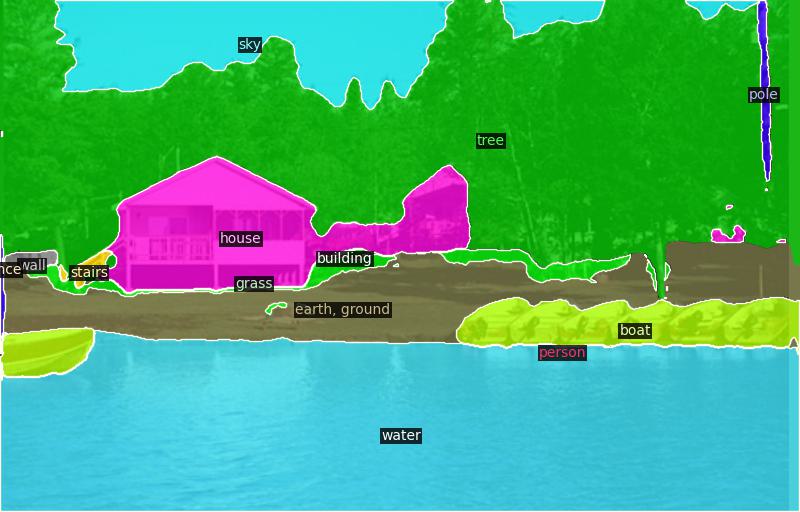} 
   \end{subfigure}
   \hfill
   \begin{subfigure}{0.32\linewidth}
    \includegraphics[width=1.0\columnwidth]{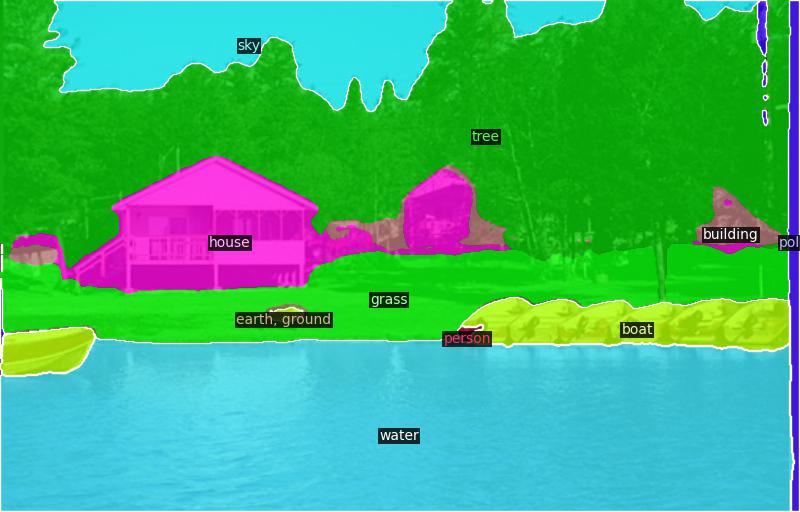} 
   \end{subfigure}

   \begin{subfigure}{0.32\linewidth}
        \includegraphics[width=1.0\columnwidth]{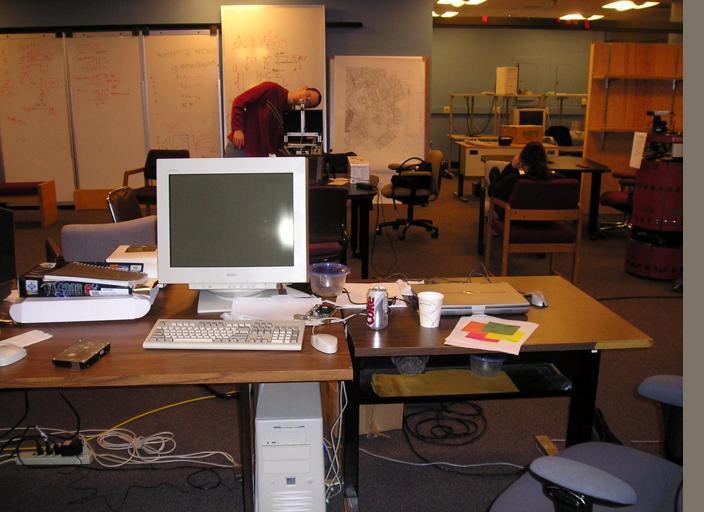} 
   \end{subfigure}
   \hfill
   \begin{subfigure}{0.32\linewidth}
    \includegraphics[width=1.0\columnwidth]{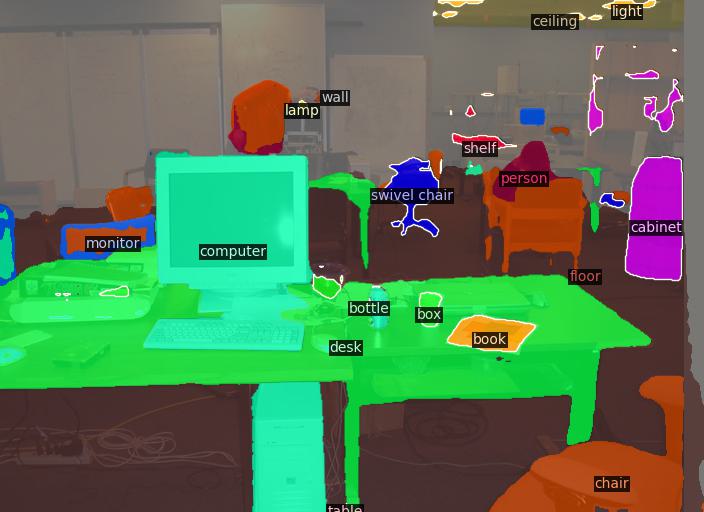} 
   \end{subfigure}
   \hfill
   \begin{subfigure}{0.32\linewidth}
    \includegraphics[width=1.0\columnwidth]{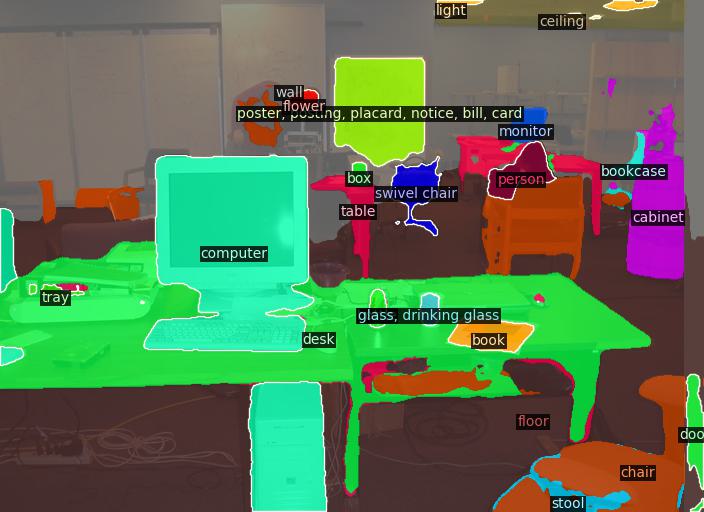} 
   \end{subfigure}

   \begin{subfigure}{0.32\linewidth}
        \includegraphics[width=1.0\columnwidth]{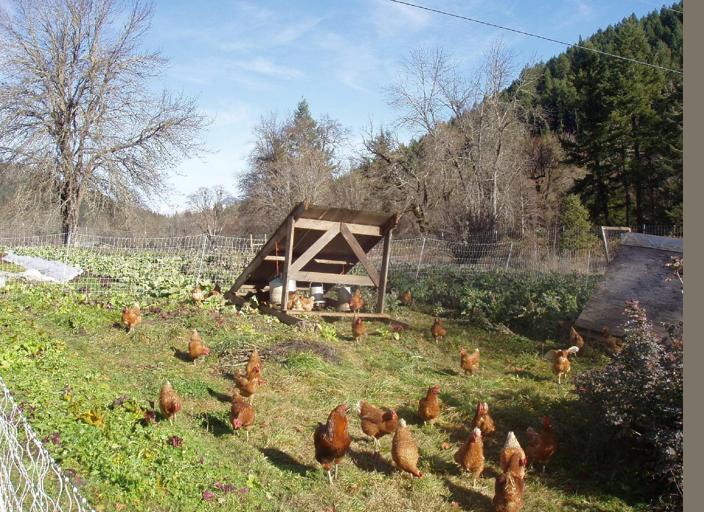} 
   \end{subfigure}
   \hfill
   \begin{subfigure}{0.32\linewidth}
    \includegraphics[width=1.0\columnwidth]{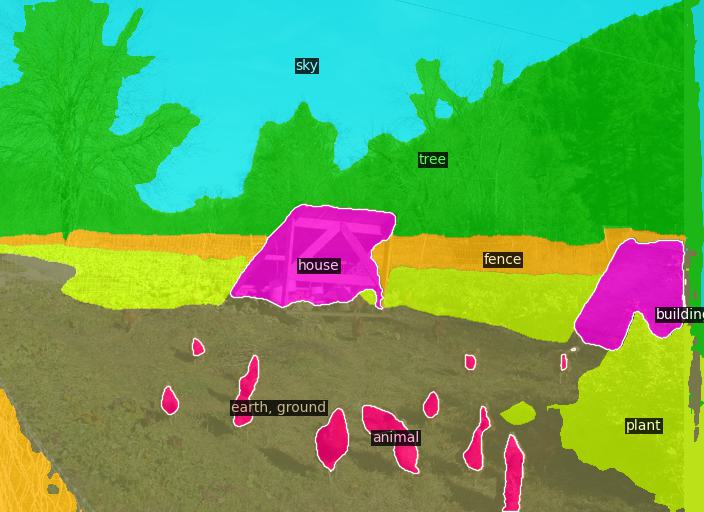} 
   \end{subfigure}
   \hfill
   \begin{subfigure}{0.32\linewidth}
    \includegraphics[width=1.0\columnwidth]{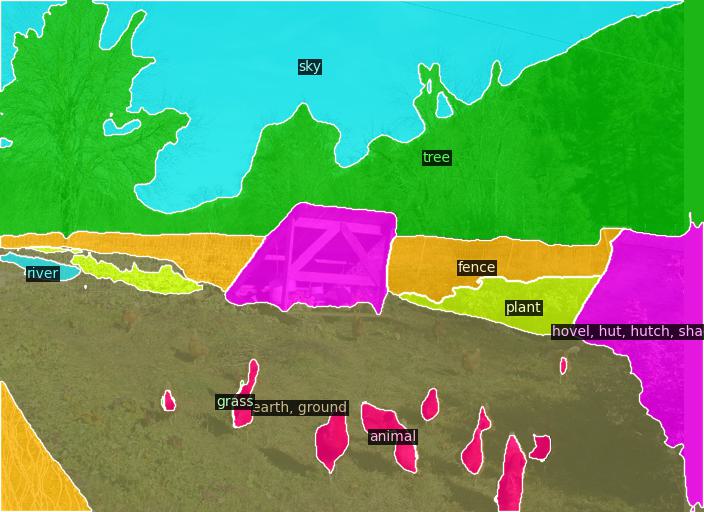} 
   \end{subfigure}

   \begin{subfigure}{0.32\linewidth}
        \includegraphics[width=1.0\columnwidth]{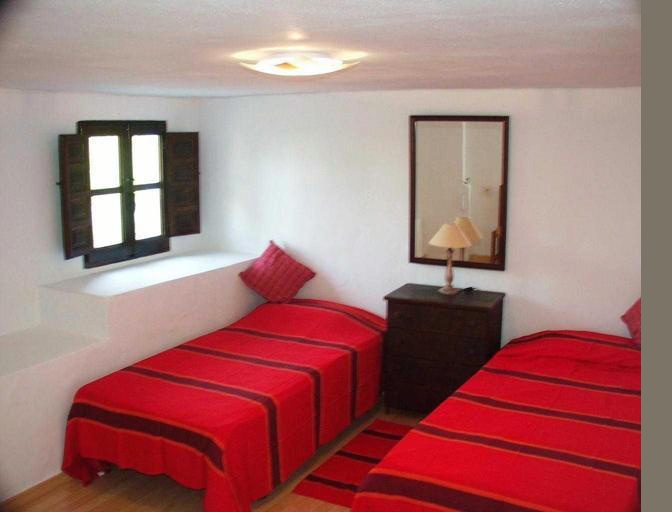} 
     \caption{Original images.}
   \end{subfigure}
   \hfill
   \begin{subfigure}{0.32\linewidth}
    \includegraphics[width=1.0\columnwidth]{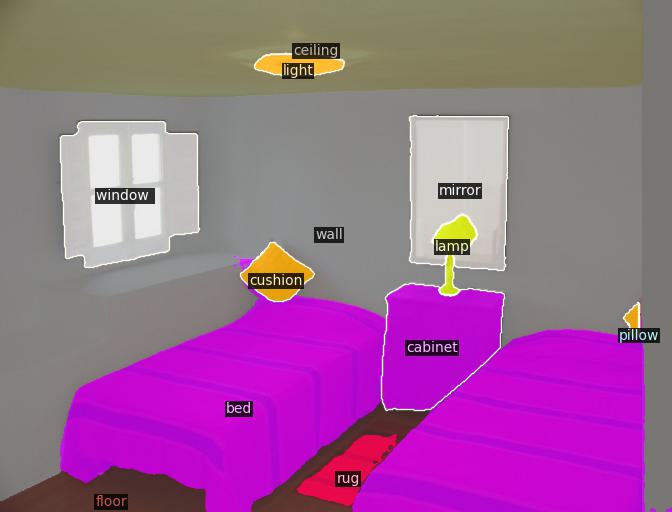} 
     \caption{AFF-Tiny predictions.}
   \end{subfigure}
   \hfill
   \begin{subfigure}{0.32\linewidth}
    \includegraphics[width=1.0\columnwidth]{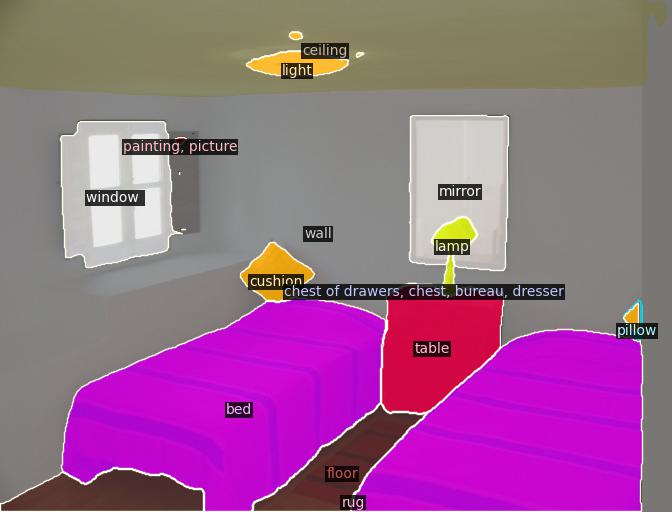} 
     \caption{Swin-Tiny prediction.}
   \end{subfigure}
    \caption{Qualitative comparison between AFF-Tiny and Swin-Tiny with Mask2Former segmentation head on ADE20K semantic segmentation. First column: original image. Second column: AFF prediction. Third column: Swin prediction.}
    \label{fig:examples_ade}
\end{figure*}

In Fig.~\ref{fig:examples_ade}, we provide a qualitative comparison between AFF-Tiny and Swin-Tiny with Mask2Former segmentation head on the ADE20K semantic segmentation data. Our model  better  captures small objects (e.g., the pole in the first row, the chickens in the third row, and the rug in the fourth row) with fewer false positives in small objects (compared to the Swin baseline in the second row).

\begin{figure*}
    \centering
    \begin{footnotesize}
    \begin{tabular}{c@{ }c@{ }c@{ }c@{ }c@{ }c}
       Image & Stage 1 & Stage 2 & Stage 3 & Stage 4 & \makecell{Clustering\\Configuration}\\
    \begin{subfigure}{0.16\linewidth}
        \includegraphics[width=1.0\columnwidth]{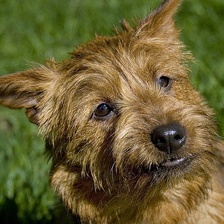} 
   \end{subfigure} &
   \begin{subfigure}{0.16\linewidth}
    \includegraphics[width=1.0\columnwidth]{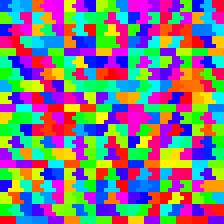} 
   \end{subfigure} &
   \begin{subfigure}{0.16\linewidth}
    \includegraphics[width=1.0\columnwidth]{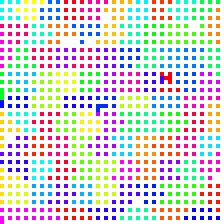} 
   \end{subfigure} &
   \begin{subfigure}{0.16\linewidth}
    \includegraphics[width=1.0\columnwidth]{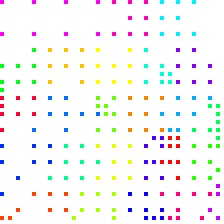} 
   \end{subfigure} &
   \begin{subfigure}{0.16\linewidth}
    \includegraphics[width=1.0\columnwidth]{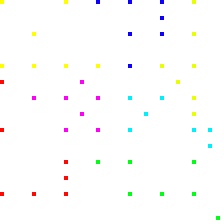} 
   \end{subfigure} & 
   \begin{subfigure}{0.16\linewidth}
   \begin{tikzpicture}
    \node[rectangle,fill=white,minimum width = 3cm, 
    minimum height = 3cm] (r) at (0,0) {
    \makecell{With anchors;\\Horizontal Scanline}
    };
    \end{tikzpicture}
   \end{subfigure} \\

    &
   \begin{subfigure}{0.16\linewidth}
    \includegraphics[width=1.0\columnwidth]{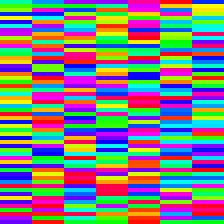} 
   \end{subfigure} &
   \begin{subfigure}{0.16\linewidth}
    \includegraphics[width=1.0\columnwidth]{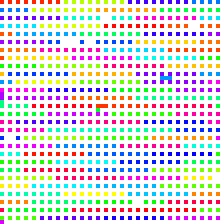} 
   \end{subfigure} &
   \begin{subfigure}{0.16\linewidth}
    \includegraphics[width=1.0\columnwidth]{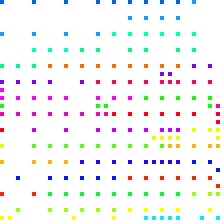} 
   \end{subfigure} &
   \begin{subfigure}{0.16\linewidth}
    \includegraphics[width=1.0\columnwidth]{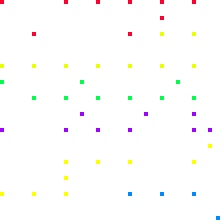} 
   \end{subfigure} & 
   \begin{subfigure}{0.16\linewidth}
   \begin{tikzpicture}
    \node[rectangle,fill=white,minimum width = 3cm, 
    minimum height = 3cm] (r) at (0,0) {
    \makecell{Without anchors;\\Horizontal Scanline}
    };
    \end{tikzpicture}
   \end{subfigure} \\
   
     &
   \begin{subfigure}{0.16\linewidth}
    \includegraphics[width=1.0\columnwidth]{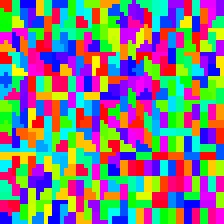} 
   \end{subfigure} &
   \begin{subfigure}{0.16\linewidth}
    \includegraphics[width=1.0\columnwidth]{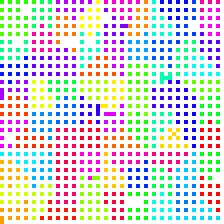} 
   \end{subfigure} &
   \begin{subfigure}{0.16\linewidth}
    \includegraphics[width=1.0\columnwidth]{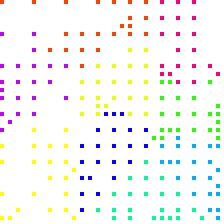} 
   \end{subfigure} &
   \begin{subfigure}{0.16\linewidth}
    \includegraphics[width=1.0\columnwidth]{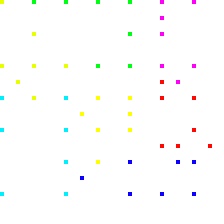} 
   \end{subfigure} & 
   \begin{subfigure}{0.16\linewidth}
   \begin{tikzpicture}
    \node[rectangle,fill=white,minimum width = 3cm, 
    minimum height = 3cm] (r) at (0,0) {
    \makecell{With anchors;\\Peano Curve}
    };
    \end{tikzpicture}
   \end{subfigure} \\

   &
   \begin{subfigure}{0.16\linewidth}
    \includegraphics[width=1.0\columnwidth]{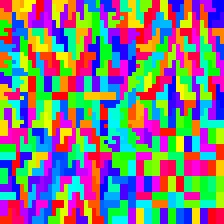} 
   \end{subfigure} &
   \begin{subfigure}{0.16\linewidth}
    \includegraphics[width=1.0\columnwidth]{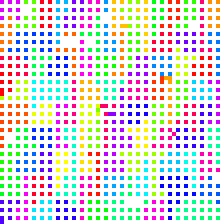} 
   \end{subfigure} &
   \begin{subfigure}{0.16\linewidth}
    \includegraphics[width=1.0\columnwidth]{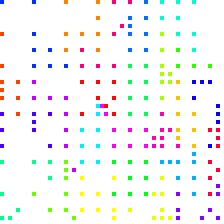} 
   \end{subfigure} &
   \begin{subfigure}{0.16\linewidth}
    \includegraphics[width=1.0\columnwidth]{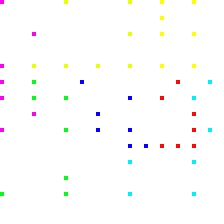} 
   \end{subfigure} & 
   \begin{subfigure}{0.16\linewidth}
   \begin{tikzpicture}
    \node[rectangle,fill=white,minimum width = 3cm, 
    minimum height = 3cm] (r) at (0,0) {
    \makecell{Without anchors;\\Peano Curve}
    };
    \end{tikzpicture}
   \end{subfigure} \\

   &
   \begin{subfigure}{0.16\linewidth}
    \includegraphics[width=1.0\columnwidth]{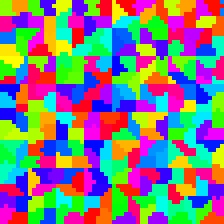} 
   \end{subfigure} &
   \begin{subfigure}{0.16\linewidth}
    \includegraphics[width=1.0\columnwidth]{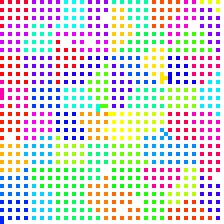} 
   \end{subfigure} &
   \begin{subfigure}{0.16\linewidth}
    \includegraphics[width=1.0\columnwidth]{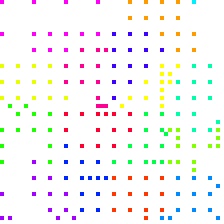} 
   \end{subfigure} &
   \begin{subfigure}{0.16\linewidth}
    \includegraphics[width=1.0\columnwidth]{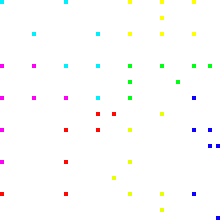} 
   \end{subfigure} & 
   \begin{subfigure}{0.16\linewidth}
   \begin{tikzpicture}
    \node[rectangle,fill=white,minimum width = 3cm, 
    minimum height = 3cm] (r) at (0,0) {
    \makecell{With anchors;\\Hilbert Curve}
    };
    \end{tikzpicture}
   \end{subfigure} \\
   
   &
   \begin{subfigure}{0.16\linewidth}
    \includegraphics[width=1.0\columnwidth]{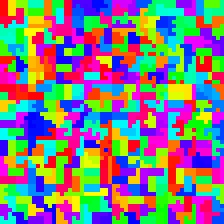} 
   \end{subfigure} &
   \begin{subfigure}{0.16\linewidth}
    \includegraphics[width=1.0\columnwidth]{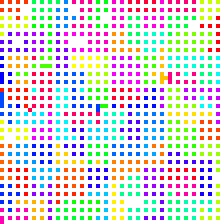} 
   \end{subfigure} &
   \begin{subfigure}{0.16\linewidth}
    \includegraphics[width=1.0\columnwidth]{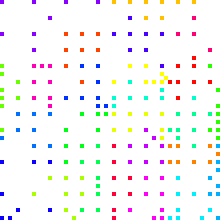} 
   \end{subfigure} &
   \begin{subfigure}{0.16\linewidth}
    \includegraphics[width=1.0\columnwidth]{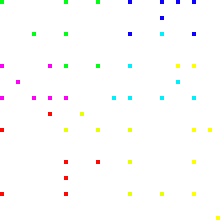} 
   \end{subfigure} & 
   \begin{subfigure}{0.16\linewidth}
   \begin{tikzpicture}
    \node[rectangle,fill=white,minimum width = 3cm, 
    minimum height = 3cm] (r) at (0,0) {
   \makecell{Without anchors;\\Hilbert Curve}
    };
    \end{tikzpicture}
   \end{subfigure} \\
    \end{tabular}
    \end{footnotesize}
    \caption{Visualization of the balanced clustering results with different configurations of anchors and space-filling curves. For the cases without anchors, the space-filling curve is applied directly on the tokens. From the results, we observe the use of anchors to be critical for obtaining more rounded and separated clusters. Although Peano and Hilbert are recursive curves, the uneven density of the tokens due to adaptive sampling still breaks the local euclidean metric if we directly apply these curves on the tokens.}
    \label{fig:cluster}
\end{figure*}

\begin{figure*}
    \centering
    \begin{footnotesize}
    \begin{tabular}{c}
\vspace{1em}
   \begin{subfigure}{\linewidth}
    \includegraphics[width=1.0\columnwidth]{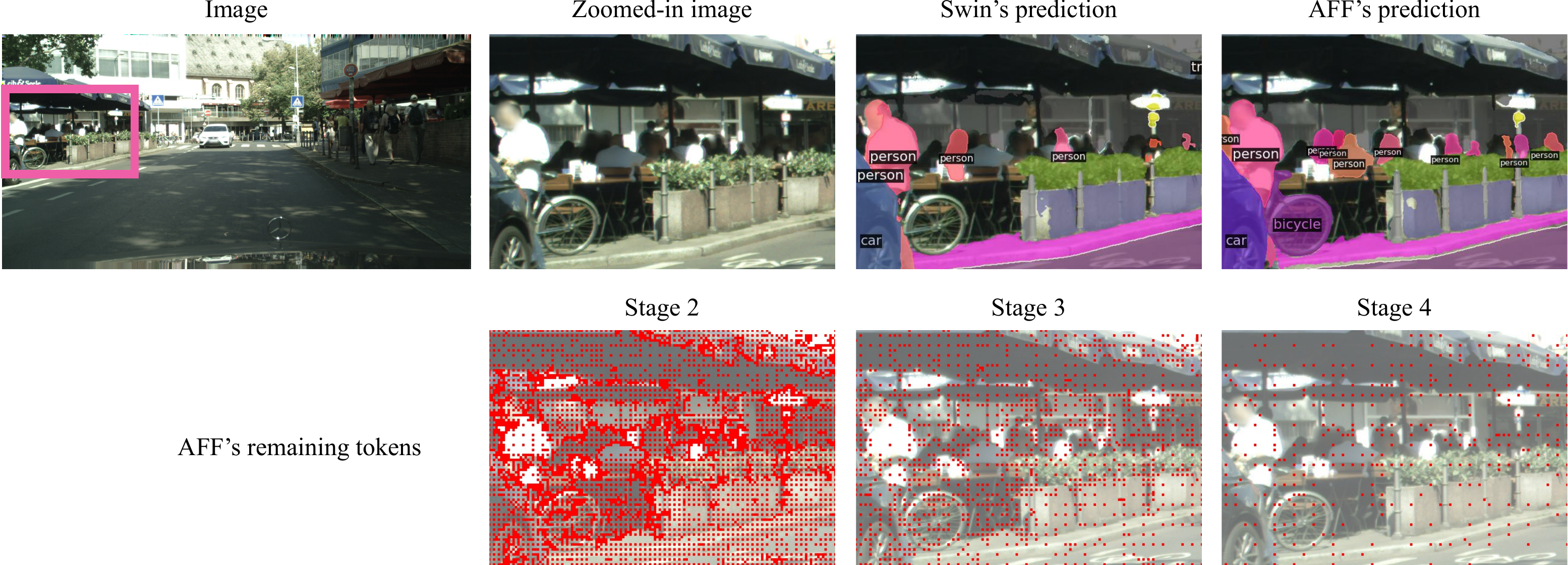} 
   \end{subfigure} \\
\vspace{1em}
   \begin{subfigure}{\linewidth}
    \includegraphics[width=1.0\columnwidth]{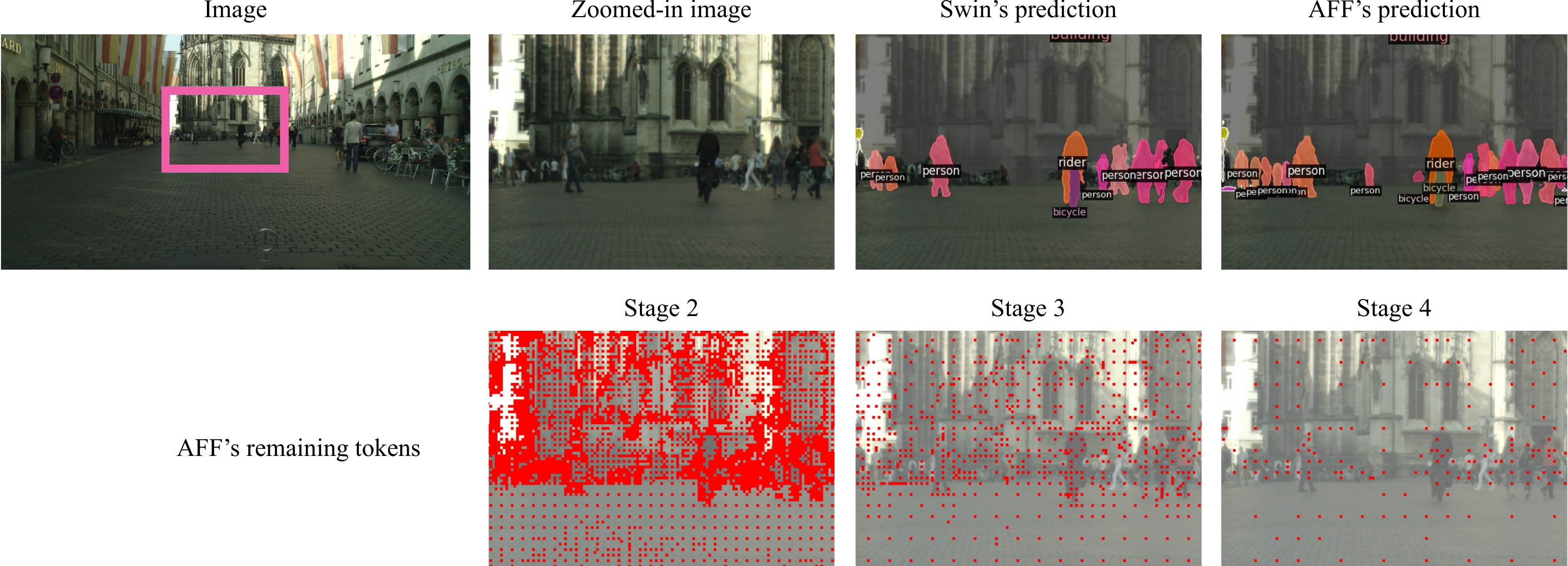} 
   \end{subfigure} \\
\vspace{1em}
   \begin{subfigure}{\linewidth}
    \includegraphics[width=1.0\columnwidth]{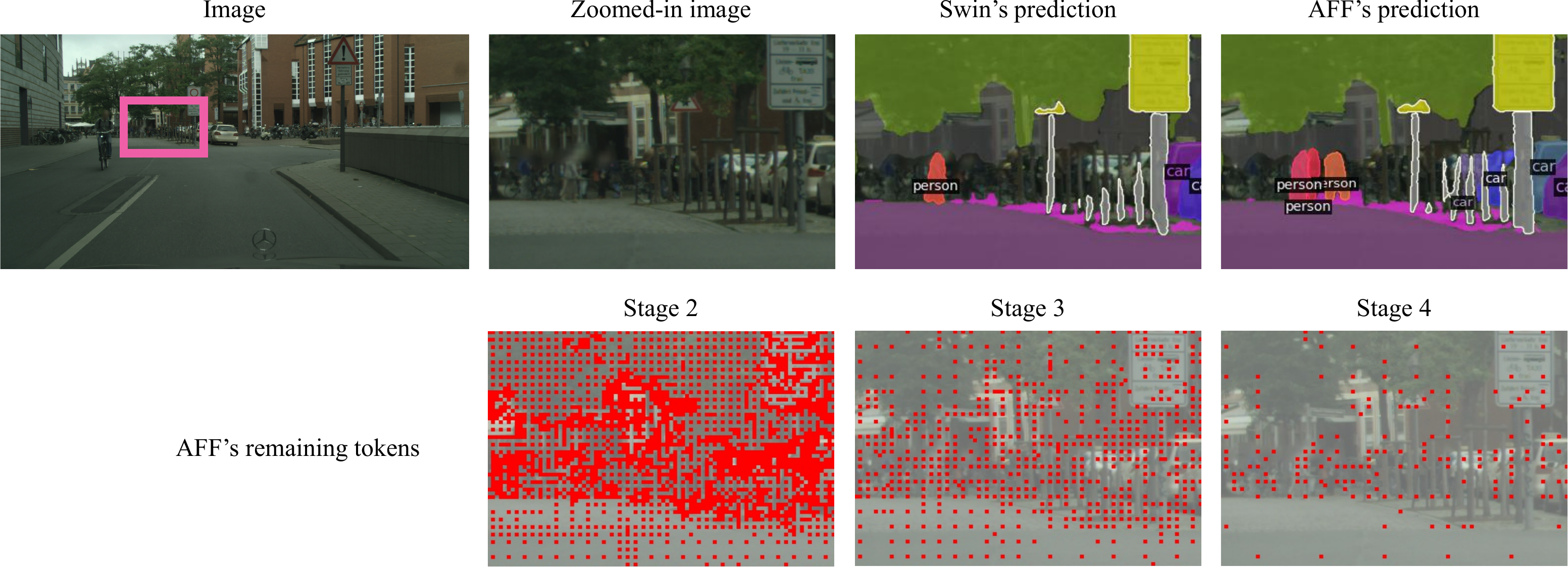} 
   \end{subfigure} \\
   
   \end{tabular}
    \end{footnotesize}
    \caption{Additional qualitative comparison between AFF-Small and Swin-Small with Mask2Former segmentation head on Cityscapes panoptic segmentation. The red pixels in the even rows indicate the locations of the remaining tokens in stage 2, 3 and 4.}
    \label{fig:city}
\end{figure*}

\begin{figure*}\ContinuedFloat
    \centering
    \begin{footnotesize}
    \begin{tabular}{c}

\vspace{1em}
   \begin{subfigure}{\linewidth}
    \includegraphics[width=1.0\columnwidth]{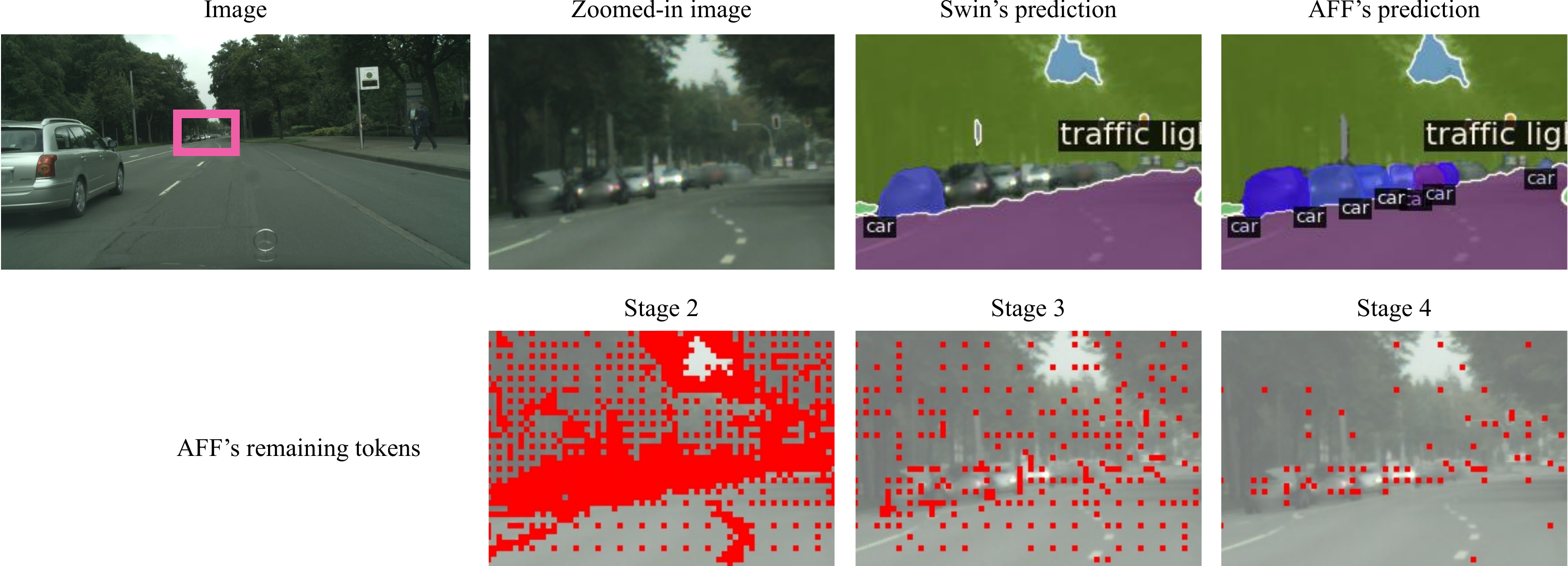} 
   \end{subfigure} \\

\vspace{1em}
   \begin{subfigure}{\linewidth}
    \includegraphics[width=1.0\columnwidth]{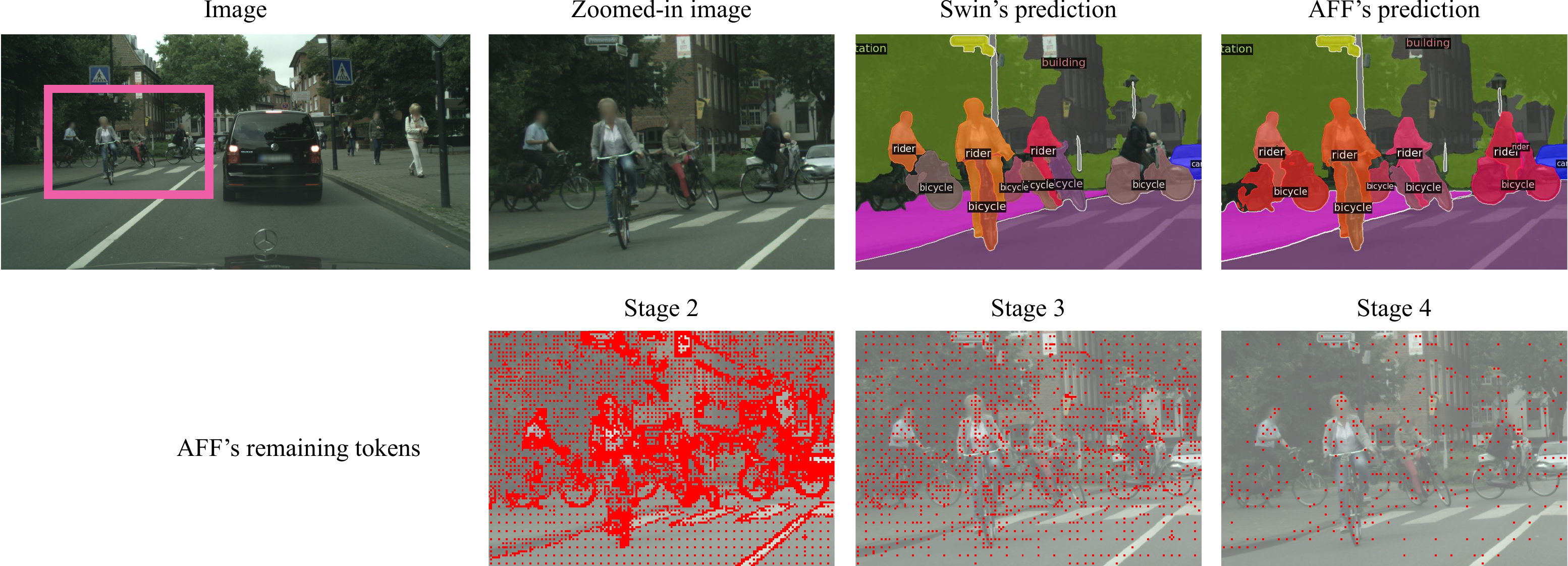} 
   \end{subfigure} \\
\vspace{1em}
   \begin{subfigure}{\linewidth}
    \includegraphics[width=1.0\columnwidth]{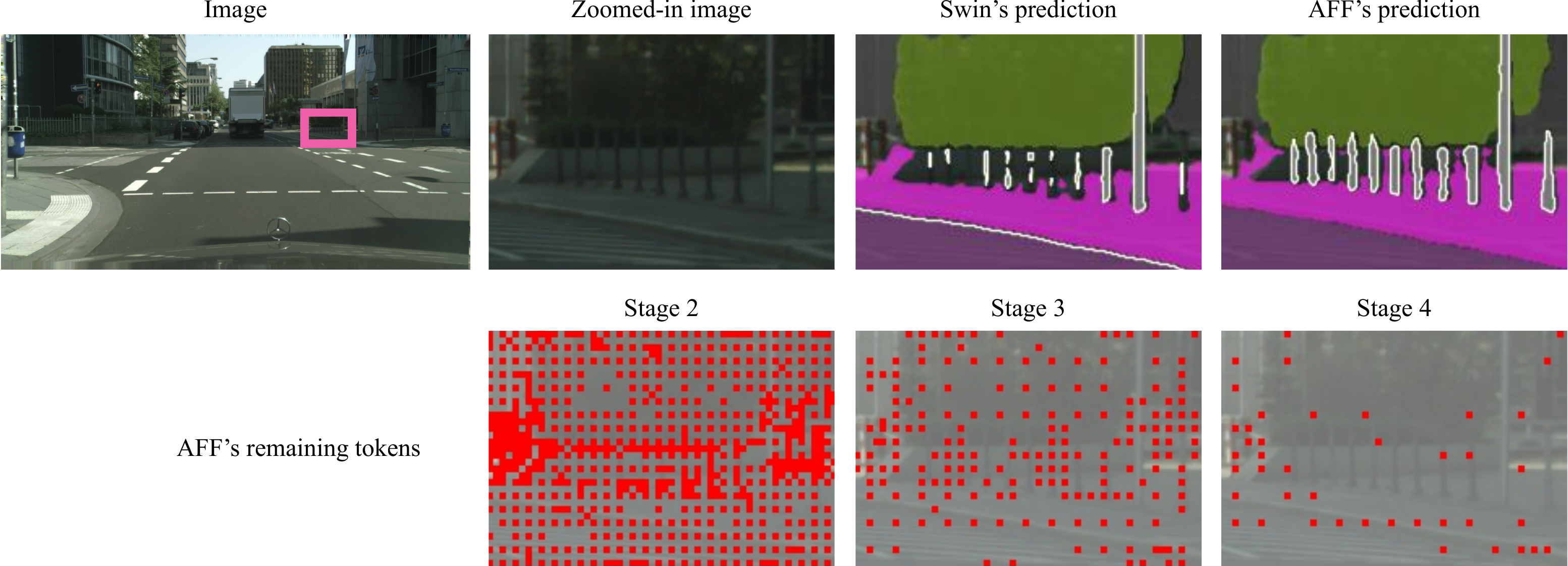} 
   \end{subfigure} \\

       \end{tabular}
    \end{footnotesize}
    \caption{(Continued) Additional qualitative comparison between AFF-Small and Swin-Small with Mask2Former segmentation head on Cityscapes panoptic segmentation. The red pixels in the even rows indicate the locations of the remaining tokens in stage 2, 3 and 4.}
\end{figure*}

\end{document}